\documentclass[12pt,authoryear]{elsarticle}
\usepackage[utf8]{inputenc}
\usepackage{xcolor}
\usepackage{hyperref}
\usepackage{amsmath}
\usepackage{natbib}
\usepackage{multirow}
\usepackage{graphicx}
\usepackage{tabularx}
\usepackage{makecell}
\usepackage{siunitx}
\usepackage{caption}
\usepackage{subcaption}
\usepackage{ragged2e}
\usepackage{float}
\usepackage[normalem]{ulem}
\usepackage{orcidlink}
\usepackage{algorithm}
\usepackage[noend]{algpseudocode}
\usepackage{setspace}
\usepackage{longtable}
\makeatletter
\def\ps@pprintTitle{%
  \let\@oddhead\@empty
  \let\@evenhead\@empty
  \def\@oddfoot{\reset@font\hfil\thepage\hfil}
  \let\@evenfoot\@oddfoot
}
\makeatother

\begin{document}

\begin{frontmatter}
    %% Title
    \title{Standardization for improved Spatio-Temporal Image Fusion}
    %% Authors
    \author[1,2]{Harkaitz Goyena \orcidlink{0000-0003-1963-9245}}
    \author[3]{Peter M. Atkinson}
    \author[1,2]{Unai P\'erez-Goya \orcidlink{0000-0002-2796-9079}} 
    \author[1,2]{M. Dolores Ugarte \orcidlink{0000-0002-3505-8400}}
    \address[1]{Department of Statistics, Computer Science and Mathematics, Public University of Navarre, Pamplona, Spain}
    \address[2]{InaMat$^2$ Institute, Pamplona, Spain}
    \address[3]{Lancaster Environment Centre, Lancaster University, Bailrigg Lancaster LA1 4YW, UK}
    
    \begin{abstract}
        Spatio-Temporal Image Fusion (STIF) methods usually require sets of images with matching spatial and spectral resolutions captured by different sensors. To facilitate the application of STIF methods, we propose and compare two different standardization approaches. The first method is based on traditional upscaling of the fine-resolution images. The second method is a sharpening approach called Anomaly Based Satellite Image Standardization (ABSIS) that blends the overall features found in the fine-resolution image series with the distinctive attributes of a specific coarse-resolution image to produce images that more closely resemble the outcome of aggregating the fine-resolution images. Both methods produce a significant increase in accuracy of the Unpaired Spatio Temporal Fusion of Image Patches (USTFIP) STIF method, with the sharpening approach increasing the spectral and spatial accuracies of the fused images by up to $49.46\%$ and $78.40\%$, respectively.
    \end{abstract}
    
    \begin{keyword}
        Satellite sensor imagery \sep spatio-temporal image fusion \sep upscaling
    \end{keyword}
    
\end{frontmatter}

\section{Introduction} \label{sec:intro}

%% relevance
Satellite sensor images are a powerful tool for Earth observation due to the vast number of sensor constellations in orbit. These sensors provide valuable information in many research areas such as ecology \citep{ecology2018}, agriculture \citep{agro2020}, urban studies \citep{urban2020}, and economics \citep{economic2016}. Many studies rely on satellite images from publicly available sources like the Landsat \citep{landsat7, landsat8}, MODIS \citep{modis} and Sentinel programs \citep{sentinel2, sentinel3}. However, each satellite program needs to make a compromise between the sensor spatial and temporal resolutions due to the trade-offs between swath-width and revisiting time. For example, Sentinel-$2$ and Sentinel-$3$ scan the Earth's surface every $10$ and $2$ days at $10$ m-or-$20$ m and $300$ m resolution, respectively \citep{sentinel2,sentinel3}. Spatio-Temporal Image Fusion (STIF) methods circumvent this issue by blending images from complementary sources to obtain the ``best of both worlds'' in terms of spatial and temporal resolutions \citep{ghamisi2019}.

%% STIF overview
The high demand for more spatially detailed images and the complexity of the problem being addressed by STIF has resulted in an abundance of techniques, generally grouped into three major categories depending on the principles applied during the fusion process \citep{chen2015, belgiu2019}: reconstruction-based, unmixing-based, and learning-based methods. Still, these methods are usually highly sensitive to discrepancies between sensors \citep{coreg2015,coreg2020,coreg2020patch,zhou2021}. Thus, standardizing remote sensing images is crucial for ensuring compatibility and interoperability across diverse datasets. However, challenges such as spectral normalization, addressing variations in sensor characteristics, atmospheric conditions and acquisition angles, and spatial pattern fitting to manage differences in spatial resolution and geometric distortions complicate this process. Advanced correction techniques, including radiometric normalization and geometric co-registration, are essential to mitigate these discrepancies and ensure consistent and precise information extraction from remote sensing imagery \citep{heo2000standardized, abdelrahman2011solving}. Given these challenges, a robust standardization approach is essential to maximize the usability of STIF methods. Several techniques have been developed to improve input data quality based on different fundamental approaches. One approach consists of correcting radiometric discrepancies, which requires somewhat correlated images captured by different sensors. This problem is usually solved by the use of pseudo-invariant areas \citep{pons2014automatic}. Another approach consists of aligning the images to correct spatial discrepancies \citep{sommervold2023survey}. For example, \cite{emelyanova2013} shifted the images horizontally and vertically to rectify estimation errors arising from the resampling process. Since most of these methods involve feature extraction and matching, deep learning methods are gaining interest in this field \citep{wu2021computational}. Despite their usefulness in certain cases, each approach has its pitfalls: image shifting requires a matching pair of images to provide the standardized image, while the other approaches require vast numbers of data, either to find the pseudo-invariant areas or to train the feature extraction or matching. 

Most STIF methods address the standardization or image-pairing problem primarily through co-registration \citep{belgiu2019}, often by utilizing previously co-registered image pairs. Other challenges related to multi-sensor data are typically handled within the framework of the specific fusion model, frequently relying on more complex approaches, such as deep learning techniques, to handle nonlinear relationships \citep{vivone2025DLreview}. While this research focuses on improving the input data for STIF methods, it takes a broader approach by addressing multi-sensor challenges at their root, using generalizable methods applicable to a wide range of multi-sensor remote sensing tasks.

%% Proposal
To address the standardization problem, and assess its effect on STIF, we propose two different standardization approaches. The first method involves an automatic combination of the Point Spread Function (PSF) and image co-registration to improve similarity when upscaling the fine-resolution data so they resemble the coarse-resolution images. The second method sharpens a specific coarse-resolution image by blending the overall features found in the fine-resolution image series with the distinctive attributes of the specific coarse-resolution image. This technique aims to produce images that more closely resemble the outcome of resampling the fine-resolution images, thereby enhancing the overall results of STIF methods that work by transferring the coarse resolution change into the fine resolution, such as Fit-FC \citep{fitfc2018} or Unpaired Spatio-Temporal Fusion of Image Patches (USTFIP) \citep{goyena2023ustfip}. This approach ensures that fine-scale spatial patterns are better preserved, while maintaining consistency across temporal observations, which allows for improved transfer of the captured change.

The rest of the paper is organized as follows: Section \ref{sec:std_methods} introduces the proposed methodological approaches. Section \ref{sec:experiments} describes the experimental set-up and reports the results, while Section \ref{sec:discussion} discusses the proposed approaches. Concluding remarks are given in Section \ref{sec:conclusion}.

\section{Standardization approaches} \label{sec:std_methods}

To address the challenges associated with inconsistencies in satellite imagery, we introduce two distinct methodological approaches to tackle standardization. Section \ref{sec:ups} describes an upscaling-based approach that combines simulation of the Point Spread Function (PSF) and automatic co-registration to provide upscaled images that resemble the coarse-resolution images. Section \ref{sec:ABSIS} describes a sharpening approach, called Anomaly Based Satellite Image Standardization (ABSIS). Both approaches are designed to standardize level-2 processed satellite images, that is, georeferenced and atmospherically corrected remote sensing data that provide surface reflectance values essential for accurate multi-temporal and multi-sensor analysis. These images often exhibit spectral and spatial discrepancies due to sensor differences, varying atmospheric conditions and misalignments, thus requiring a robust standardization approach.

To describe both methodologies formally, we denote the set of coarse-resolution images by $C_\mathbf{T}(\mathbf{S},\mathbf{b})$, the set of fine-resolutions images as $F_\mathbf{T}(\mathbf{s},\mathbf{b})$ and the aggregated fine-resolution images as $F_\mathbf{T}(\mathbf{S}_F,\mathbf{b})$, where $\mathbf{S} = \{S_1,\ldots,$ $S_i,\ldots,S_{n}\}$ and $\mathbf{s} = \{s_1,\ldots,s_i,\ldots,s_N\}$ are the sets of pixel locations that contain measurements made on a coarse- and fine-resolution support, respectively, and $\mathbf{S}_F = \{S_{1_F},\ldots,S_i,\ldots,S_{n_F}\}$ are the aggregated fine-resolution support pixel locations. We denote the capture dates as $\mathbf{T} = \{t_1,\ldots,t_k,$ $\ldots,t_{K}\}$ and $\mathbf{b} = \{b_1,\ldots,b_l,\ldots,b_L\}$ denotes the set of multispectral bands contained in each image. Therefore, $C_{t_k}(S_i,b_l)$ represents the reflectance value of the coarse image at date $t_k$ for band $b_l$ at the coarse-resolution support pixel location $S_i$.

\subsection{Automatic Upscaling} \label{sec:ups}

Upscaling fine-resolution images is typically the preferred approach for enhancing the similarity between coarse and fine-resolution satellite images, as it is generally easier to degrade image quality than to recover detail from coarse-resolution data.

\subsubsection{Simulate Sentinel-3 Point Spread Function}

\begin{figure} [!h]
    \includegraphics[width=\linewidth]{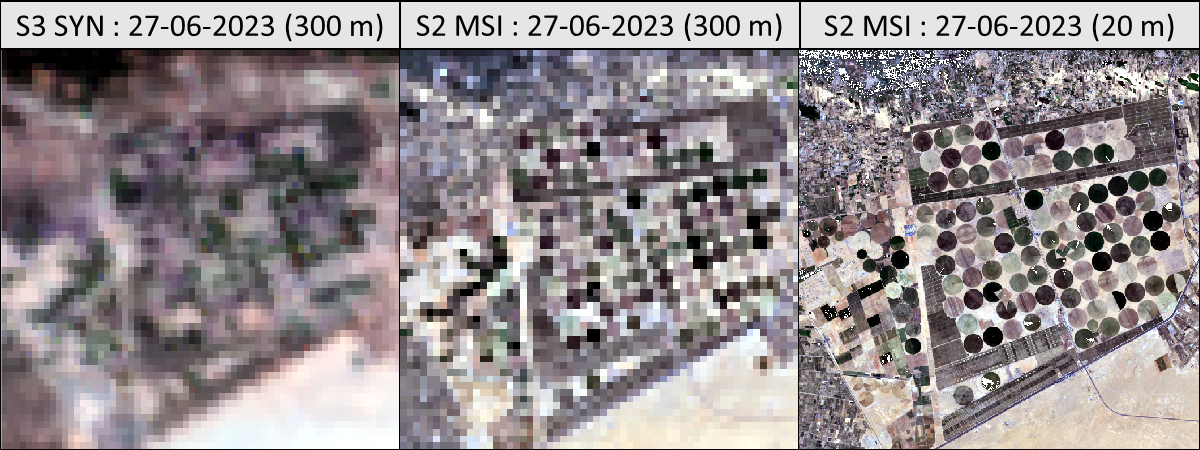}
    \caption{RGB representations, from left to right, of the smoothed Sentinel-3 SYN image, aggregated Sentinel-2 and original Sentinel-2 MSI image for the croplands region.}
    \label{fig:mot_ups_cl}
\end{figure}

\begin{figure} [!h]
    \includegraphics[width=\linewidth]{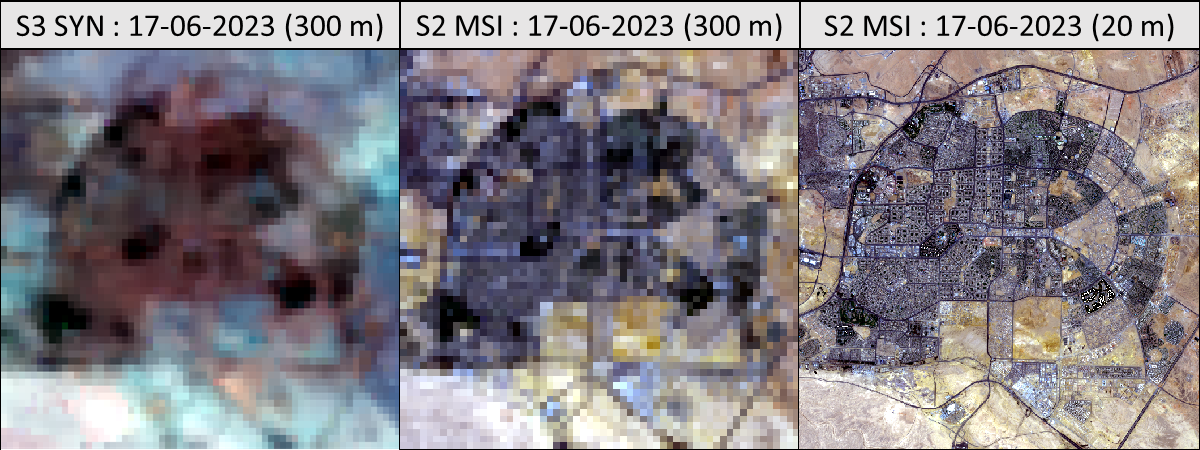}
    \caption{RGB representations, from left to right, of the smoothed Sentinel-3 SYN image, aggregated Sentinel-2 and original Sentinel-2 MSI image for the New Cairo region.}
    \label{fig:mot_ups_nc}
\end{figure}

Sentinel-3 images appear significantly more blurry compared to their matching Sentinel-2 images (see Figures \ref{fig:mot_ups_cl} and \ref{fig:mot_ups_nc}), suggesting that direct bilinear interpolation or averaging of Sentinel-2 data may be insufficient. The aggregated image for the Croplands region (Figure \ref{fig:mot_ups_cl}) even shows some signs of aliasing when sampling the circle-like forms, which is not present in the Sentinel-3 imagery. To address this issue, we begin by incorporating a simulation of the Sentinel-3 Point Spread Function (PSF) during the upscaling process of Sentinel-2 data.

We approximate the Sentinel-3 PSF by a Gaussian filter, with a standard deviation $\sigma$ that maximizes the linear correlation between the upscaled Sentinel-2 and the corresponding Sentinel-3 images. The upscaled Sentinel-2 image, denoted by $\hat{C}_{t_k}$, is computed by convolving the Sentinel-2 image $F_{t_k}$ with a Gaussian kernel
\begin{equation}
    \hat{C}_{t_k}(\mathbf{S},b) = F_{t_k}(\mathbf{s},b) * g_{\sigma}, \quad t_k \in \mathbf{T}
\end{equation}
where $g_{\sigma}$ is a Gaussian filter with standard deviation $\sigma$, written as:
\begin{equation}\label{eq:GF}
    g_{\sigma}(d) = \frac{1}{2 \pi \sigma^2} \text{exp}(-\frac{d^2}{2\sigma^2}),
\end{equation}
where $d$ is the Euclidean distance from the center of each Sentinel-2 pixel to the Sentinel-3 pixel center. 

The optimal standard deviation $\sigma$ for the Gaussian filter is selected individually for each spectral band and image pair as the value that maximizes the Pearson’s correlation coefficient between the Sentinel-3 image and the upscaled Sentinel-2 image. The search space for $\sigma$ consists of values ranging from 0.4 to 2 theoretical Sentinel-3 pixels, in increments of 0.1.
For reference, $\sigma=1$ corresponds to a standard deviation of $300$ m.

\subsubsection{Co-registration}

The next step consists of solving the misregistration problems that came to light after applying the optimal simulated PSF. To this end, we shift the optimal Gaussian kernels obtained from the previous step to maximize the Pearson's correlation coefficient.
%The upscaled image pixel values obtained by shifting the kernel can be written as
% \begin{equation}
%     \hat{C}_{t_k,x,y}(S_i) = \sum_{s \in \mathcal{N}_{x,y}(S_i)} W(s) F_{t_k}(s), \quad S_i \in \mathbf{S}, \quad t_k \in \mathbf{T}
% \end{equation}
% where $\mathcal{N}_{x,y}(S_i)$ is a neighborhood centered around the pixel located at a distance of $x$ horizontal and $y$ vertical fine-resolution pixels from the target pixel $S_i$.
The filter weights remain the same as in the previous step, the only difference being the pixels they are applied to. The optimal shifts are selected using a grid search between $-2$ and $2$ pixels, increasing by complete pixels, which is increased on demand.

\subsubsection{Joint Gaussian Filter and Co-registration}

Upon considering the optimal parameters from the initial stages of this investigation, a correlation became apparent between the estimates of the optimal shifts and the filter deviation. More specifically, greater optimal shift estimates also exhibited greater filter deviations. This suggests that the filter may be oversmoothing in some cases to compensate for the misregistration problem. Fitting all the parameters jointly should help to alleviate this problem. However, jointly fitting all the hyperparameters greatly increases the parameter search space, hindering the use of a grid search. Consequently, a more computationally efficient search algorithm is required to optimize the hyperparameters in a timely manner. We chose a greedy algorithm as a compromise between computational cost and upscaling accuracy. For each search iteration, we increase each hyperparameter separately and choose the direction of maximal increase in the Pearson's correlation coefficient. This search algorithm corresponds to lines 1-20 of Algorithm \ref{alg:greedy}.

\subsubsection{Sub-pixel level co-registration} \label{sec:1v1ups}

Further refining the upscaling requires considering not only pixel level shifts for the required co-registration, but also sub-pixel level shifts. Considering sub-pixel shifts has two main caveats: (1) It requires to interpolate the fine-resolution pixel values at the shifted between-pixel locations, and (2) Sup-pixel level shifts require the search algorithm to explore a much larger number of hyperparameter combinations to find the optimal hyperparameters. To reduce the computational burden, the sub-pixel level shifts are searched for in a neighborhood of the pixel-level parameters, that is, in a neighborhood of the optimal parameters given by the previous greedy search algorithm, Algorithm \ref{alg:greedy} shows the pseudo-code for the updated search algorithm.

\subsubsection{Generalization} \label{sec:gralups}

The optimal parameter configurations were initially found to vary between the different dates. This arose because applying the pairwise optimal upscaling involves applying a different upscaling to each of the Sentinel-2 images, which may generate a compromise between the pairwise similarities and the time-series level similarities. Estimating generalized parameters should help to solve this problem, while also greatly increasing the computational efficiency for upscaling new Sentinel-2 images.

We estimate the generalized parameters using a modified version of the proposed greedy algorithm. Parameters are updated only if they enhance the average correlation across the available pairs. This approach balances identifying optimal upscaling parameters with maintaining the regularity of the time-series. %We use 10-fold cross-validation to assess the capability of the generalized search method.

\subsection{Anomaly-Based Satellite Image Standardization (ABSIS)} \label{sec:ABSIS}

The proposed Anomaly-Based Satellite Image Standardization (ABSIS) method is structured into three main stages. First, spatial patterns are derived from fine-resolution image time-series through pixel-wise statistical aggregation, capturing temporal trends while preserving the fine-scale spatial structure. Second, an anomaly-based correction procedure is applied to identify and adjust spectral discrepancies between the coarse and fine resolution images by modeling deviations from the expected spatial distribution. Third, the corrected anomalies are integrated with the extracted spatial patterns to generate standardized images that retain spectral consistency and spatial integrity. By systematically mitigating these discrepancies, ABSIS enhances the reliability of spatio-temporal image fusion, increasing the accuracy and consistency of satellite-derived products for applications such as environmental monitoring and land cover classification. The following sections provide a detailed description of each methodological component.

\begin{figure}
    \centering
    \includegraphics[height=0.9\textheight]{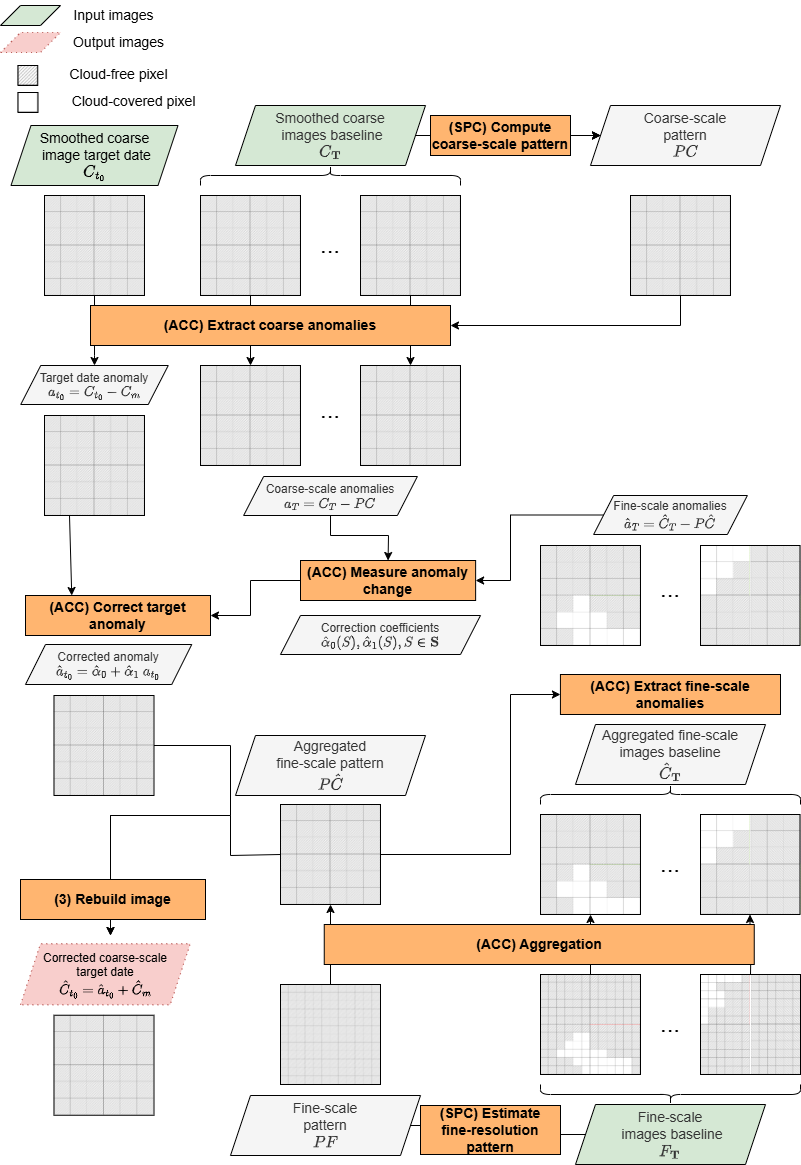}
    \caption{Flowchart of the ABSIS method}
    \label{fig:ABSIS_flowchart}
\end{figure}

Our aim is to produce images that more closely resemble the outcome of resampling the fine-resolution images. However, real images usually exhibit minimal spatial or spectral agreement, which hinders the straightforward use of linear corrections. Our method aims to maximize the utility of the spatial (pixel-by-pixel) pattern in the fine-scale images (see Section \ref{sec:SPC}) and leverage the specific behavior of the target coarse-resolution image through an anomaly-based approach (see Section \ref{sec:ACC}). Figure \ref{fig:ABSIS_flowchart} summarizes the method.

\subsubsection{Spatial Pattern Capture (SPC)} \label{sec:SPC}

At this stage, our aim is to capture the pixel-by-pixel pattern of each sensor, accomplished by using pixel-wise aggregations of the images over the baseline to obtain $P_C(\mathbf{S},\mathbf{b})$ and $P_F(\mathbf{s},\mathbf{b})$, the coarse- and fine- resolution patterns, respectively.

An straightforward and computationally efficient way to capture the spatial pattern is to compute pixel-wise means for each band. Thus, we define the coarse- and fine- resolution pixel-by-pixel patterns, $P_C(\mathbf{S}_C,\mathbf{b})$ and $P_F(\mathbf{s},\mathbf{b})$, respectively, as
\begin{equation}
    P_C(S_i, b_l) = \text{mean}(C^*_\mathbf{T}(S_i,b_l)),\ \ S_i \in \mathbf{S}, \ b_l \in \mathbf{b}
\end{equation}
and
\begin{equation}
    P_F(s_i, b_l) = \text{mean}(F^*_\mathbf{T}(s_i,b_l)),\ \ s_i \in \mathbf{s}, \ b_l \in \mathbf{b}
\end{equation}
where $C^*_\mathbf{T}(S_i,b_l)$ and $F^*_\mathbf{T}(s_i,b_l)$ are the sets of available values in \\ $\{C_{t_k}(S_i,b_l) | t_k \in \mathbf{T}\}$ and $\{ F_{t_k}(s_i,b_l) | t_k \in \mathbf{T}\}$, respectively.

\subsubsection{Anomaly Capture and Correction (ACC)} \label{sec:ACC}

The final step consists of correcting the particular behavior from image $C_{t_0}(\mathbf{S},\mathbf{b})$, represented by the anomaly $a_{t_0} = C_{t_0}(\mathbf{S},\mathbf{b}) - P_C(\mathbf{S},\mathbf{b})$. To achieve this, we transfer the anomaly change between the coarse-resolution anomalies $a_{t_k} = C_{t_k}(\mathbf{S},\mathbf{b}) - P_C(\mathbf{S},\mathbf{b}), \ t_k  \in \mathbf{T}$ and the aggregated fine-resolution anomalies $\hat{a}_{t_k} = \hat{C}_{t_k}(\mathbf{S}_F,\mathbf{b}) - P_{\hat{C}}(\mathbf{S}_F,\mathbf{b}), \ t_k  \in \mathbf{T}$, where $P_{\hat{C}}(\mathbf{S}_F,\mathbf{b})$ is the aggregated fine-resolution pattern. To be able to transfer the anomaly change, the coarse-scale anomalies need to be reprojected into the aggregated fine-resolution grid $\mathbf{S}$. We will denote the reprojected coarse-resolution anomalies for the target and baseline dates as $a'_{t_0}(\mathbf{S}_F,\mathbf{b})$ and $a'_\mathbf{T}(\mathbf{S}_F,\mathbf{b})$, respectively.

The preferred approach is to correct the coarse-resolution anomaly $a'_{t_0}$ capturing the anomaly change between the coarse- and fine-resolution images using local linear regressions between optimal pairs of observations for each pixel. We find the optimal observations using the largest correlation between $a'_{t_0}$ and the coarse-scale anomalies for the baseline $a'_{t_k},\:t_k\in\mathbf{T}$. For each pixel and band, we compute the linear correlation coefficient $\rho_{t_k}(S_i, b_l)$ as follows
\begin{equation} \label{eq:correlation}
    \rho_{t_k}(S_i, b_l) = \frac{\text{cov}(a'_{t_k}(S_i,b_l),a'_{t_0}(S_i,b_l))}{\sqrt{\text{var}(a'_{t_k}(S_i,b_l))\:\text{var}(a'_{t_0}(S_i,b_l))}}, \: t_k \in \mathbf{T}, \: S_i \in \mathbf{S}_F, \:b_l \in \mathbf{b}
\end{equation}
where $\text{cov}(a'_{t_k}(S_i,b_l),a'_{t_0}(S_i,b_l))$ is the covariance between the values inside a $w_{LR} \times w_{LR}$ window for a date $t_k$ in the baseline $\mathbf{T}$, and the values in the same window for the target date $t_0$. Both $\text{var}(\hat{C}_{t_k}(S_i,b_l))$ and $\text{var}(\hat{C}_{t_0}(S_i,b_l))$ are the variances of the values inside the same window for $t_k$ and $t_0$, respectively.

The correlation coefficient is computed only for pixels and dates with complete data, that is, when no missing values are present within the window. This allows us to capture temporal changes between the most similar windows while avoiding pixels adjacent to clouds, which are known to be error-prone. Using the results from Eq. \ref{eq:correlation}, the optimal information can be determined as follows
\begin{equation}
    \label{eq:selection}
    \delta(S_i, t_k) = \begin{cases}
        1 & \text{if $\rho_{t_k}(S_i) = \text{max}\{ \rho_\mathbf{T}(S_i) \}$}.\\
        0 & \text{otherwise},\\
    \end{cases}
\end{equation}
where $\delta$ is a mask that indicates which date $t_k$ contains the optimal observation for pixel $S_i$. Then, we can capture the change between the baseline and target time frames using the same moving window as in Eq. \ref{eq:correlation} to estimate the $\alpha_0(S_i)$ and $\alpha_1(S_i)$ coefficients in the following local linear regression
\begin{equation}
    \label{eq:lop_lr}
    a'_{t_0}(S_i) = \alpha_0(S_i)+\alpha_1(S_i) \sum_{k = 1}^{K} \delta(S_i, t_k) a'_{t_k}(S_i) + \epsilon(S_i), \quad S_i \in \mathbf{S}_F,
\end{equation}
where $\epsilon(S_i)$ is a white noise process.

Finally, we estimate the fine-resolution anomaly for the target date $\hat{a}_{t_0}(\mathbf{S}_F)$, namely
\begin{equation} \label{eq:lop_pred}
    \hat{a}_{t_0}(S_i) = \hat{\alpha}_0(S_i) + \hat{\alpha}_1(S_i)\:\sum_{k=1}^{K_F} \delta(S_i,t_k) \hat{a}_{t_k}(S_i), \quad S_i \in \mathbf{S}_F.
\end{equation}

\subsubsection{Rebuilding the corrected image}

Lastly, we combine the corrected anomaly $\hat{a}_{t_0}(\mathbf{S},\mathbf{b})$ and the aggregated fine-resolution pattern $P_{\hat{C}}(\mathbf{S},\mathbf{b})$ to obtain the corrected image at time $t_0$ as
\begin{equation}
    \hat{C}_{t_0}(S_i,b_l) = P_{\hat{C}}(S_i,b_l) + \hat{a}_{t_0}(S_i,b_l), \quad S_i \in \mathbf{S}_F, \ b_l \in \mathbf{b}.
\end{equation}

\section{Experimental Analysis} \label{sec:experiments}

We illustrate the performance of the proposed  methods through a series of experiments conducted on real satellite image datasets across two distinct geographic regions, Croplands and New Cairo, each exhibiting different spatial and temporal behaviors. First, we conduct standardization experiments to evaluate the standardization capability of each of the proposed approaches (Section \ref{sec:stdassessment}). In addition, we also perform some fusion experiments to assess the influence of the standardization methods on the results obtained by applying the USTFIP fusion method (Section \ref{sec:ustfipassessment}).

We evaluate the upscaling approach by applying both the optimal pairwise parameters (\ref{sec:1v1ups}) and the generalized parameters (\ref{sec:gralups}). To evaluate ABSIS, we simulate missing images at specific target dates and reconstruct them using a six-image baseline, consisting of three images before and three after the missing date. Since the aim of standardization is to increase similarity, we rely on the Pearson's correlation coefficient  $\rho$ for evaluation. The correlation coefficient for a given band of upscaled image $\hat{C}_{t_0}(\mathbf{S})$ is
\begin{equation}
    \rho(\hat{C}_{t_0}(\mathbf{S},b_l), {C}_{t_0}(\mathbf{S},b_l)) = \frac{\text{cov}(\hat{C}_{t_0}(\mathbf{S},b_l), {C}_{t_0}(\mathbf{S},b_l))}{\sqrt{\text{var}(\hat{C}_{t_0}(\mathbf{S},b_l))\text{var}({C}_{t_0}(\mathbf{S},b_l))}}.
\end{equation}
And the correlation for a given band of standardized image $\hat{C}_{t_0}(\mathbf{S})$ is
\begin{equation}
    \rho(\hat{C}_{t_0}(\mathbf{S}_F,b_l), {F}_{t_0}(\mathbf{S}_F,b_l)) = \frac{\text{cov}(\hat{C}_{t_0}(\mathbf{S}_F,b_l), {F}_{t_0}(\mathbf{S}_F,b_l))}{\sqrt{\text{var}(\hat{C}_{t_0}(\mathbf{S}_F,b_l))\text{var}({F}_{t_0}(\mathbf{S}_F,b_l))}},
\end{equation}
where $\text{cov}(\cdot,\cdot)$ is the sample covariance between the values from both images, and $\text{var}(\cdot)$ is the sample variance for the values of an image.

To assess the effect of the different standardization methods on the results from USTFIP, we substitute the Coarse Harmonization (CH) step from USTFIP with the different approaches and evaluate the results relying on quantitative metrics: Root Mean Square Error (RMSE) to measure spectral accuracy and Robert’s Edge feature (Edge) to assess spatial accuracy.

\noindent 
The RMSE for a given band of fused image $\hat{F}^m_{t_0}$ obtained from the standardized images generated by method $m$ is given by
\begin{equation}
    \text{RMSE}^m_{t_0}(b_l) = \sqrt{\frac{\sum_{i}^{N} (\hat{F}^m_{t_0}(s_i,b_l) - F_{t_0}(s_i,b_l))^2}{N}}, \quad b_l \in \mathbf{b}
\end{equation}
where $F_{t_0}(\mathbf{S},\mathbf{b})$ is the actual fine-resolution image, which is considered as missing during the fusion process.

\noindent
The spatial accuracy is evaluated based on the normalized difference of the Robert's Edge spatial feature, between the fused and reference images for each band $b_l$, given by
\begin{equation}
    S^{m}_{t_0}(s_i,b_l) = \frac{S_{t_0}(s_i,b_l) - \hat{S}_{t_0}^{m}(s_i,b_l)}{S_{t_0}(s_i,b_l) + \hat{S}_{t_0}^{m}(s_i,b_l)}, \quad s_i \in \mathbf{s}, \ b_l \in \mathbf{b}
\end{equation}
where $S_{t_0}$ is the spatial feature obtained from $F_{t_0}$ and $\hat{S}_{t_0}^{m}$ is the spatial feature obtained from $\hat{F}^m_{t_0}$. Then, we average the $S^{m}_{t_0}(s_i,b_l)$ values corresponding to pixels with $\hat{S}_{t_0}^{m}(s_i,b_l)$ values greater than the $90$th percentile, to obtain a single value for the spatial accuracy, denoted by $\text{Edge}^s(b_l)$. A spatial accuracy of $0$ represents a feature that is the same in both the real and the output images, negative values indicate over-smoothing in the feature for the prediction, and positive values indicate over-sharpening. For a fuller description of the Edge feature see \cite{zhu2022novel}.

RMSE quantifies deviations in reflectance values, while the Edge metric evaluates feature preservation, identifying whether predictions introduce over-smoothing or over-sharpening. The results demonstrate that both standardization approaches help enhance both spectral and spatial fidelity, with ABSIS achieving superior accuracy, as it will be shown later.

\subsection{Data} \label{sec:data}

% Regions
The experimental datasets consist of surface reflectance images cropped from scenes with cloud coverages less than $10\%$ for two different regions near Cairo, Egypt: a cropland area (Croplands) and an urban area (New Cairo). These datasets include two different multispectral satellites, Sentinel-2 and Sentinel-3, which differ in several key aspects described below.

\begin{itemize}
    \item Spectral Resolution: Sentinel-2 MSI provides finer spectral resolution with broader bands, while Sentinel-3 SYN has narrower spectral bands optimized for large-scale monitoring.  Differences in spectral resolution lead to discrepancies in band proportions between sensors, as illustrated by variations in the RGB color schemes when comparing Sentinel-2 and Sentinel-3 imagery. For example, Figure \ref{fig:mot_ups_cl} shows a slightly higher proportion of the red band in the Croplands region, whereas Figure \ref{fig:mot_ups_nc} indicates a greater presence of blue in the New Cairo region in Sentinel-3 images.
    \item Spatial Resolution: Sentinel-2 offers finer spatial resolution (20 m for the bands used), whereas Sentinel-3 operates at a coarser resolution ($\sim$300 m), hindering direct comparison between datasets due to the lack of alignment.
    \item Projection: Sentinel-2 data is delivered on a fixed grid with well-defined pixel alignment, whereas Sentinel-3 SYN data follows a non-regular grid, requiring additional interpolation and co-registration for analysis.
    \item Spatial Sampling Structure: Sentinel-2 maintains a structured, fine-resolution grid suitable for land cover analysis, while Sentinel-3’s irregular grid distribution requires spatial transformations to align with Sentinel-2 for meaningful fusion and comparison.
\end{itemize}

These differences include all the challenges we aim to address in integrating the datasets for our evaluation, ensuring the harmonization of spatial and spectral characteristics for improved spatio-temporal analysis.

\begin{figure}
    \centering
    \includegraphics[scale = 0.64]{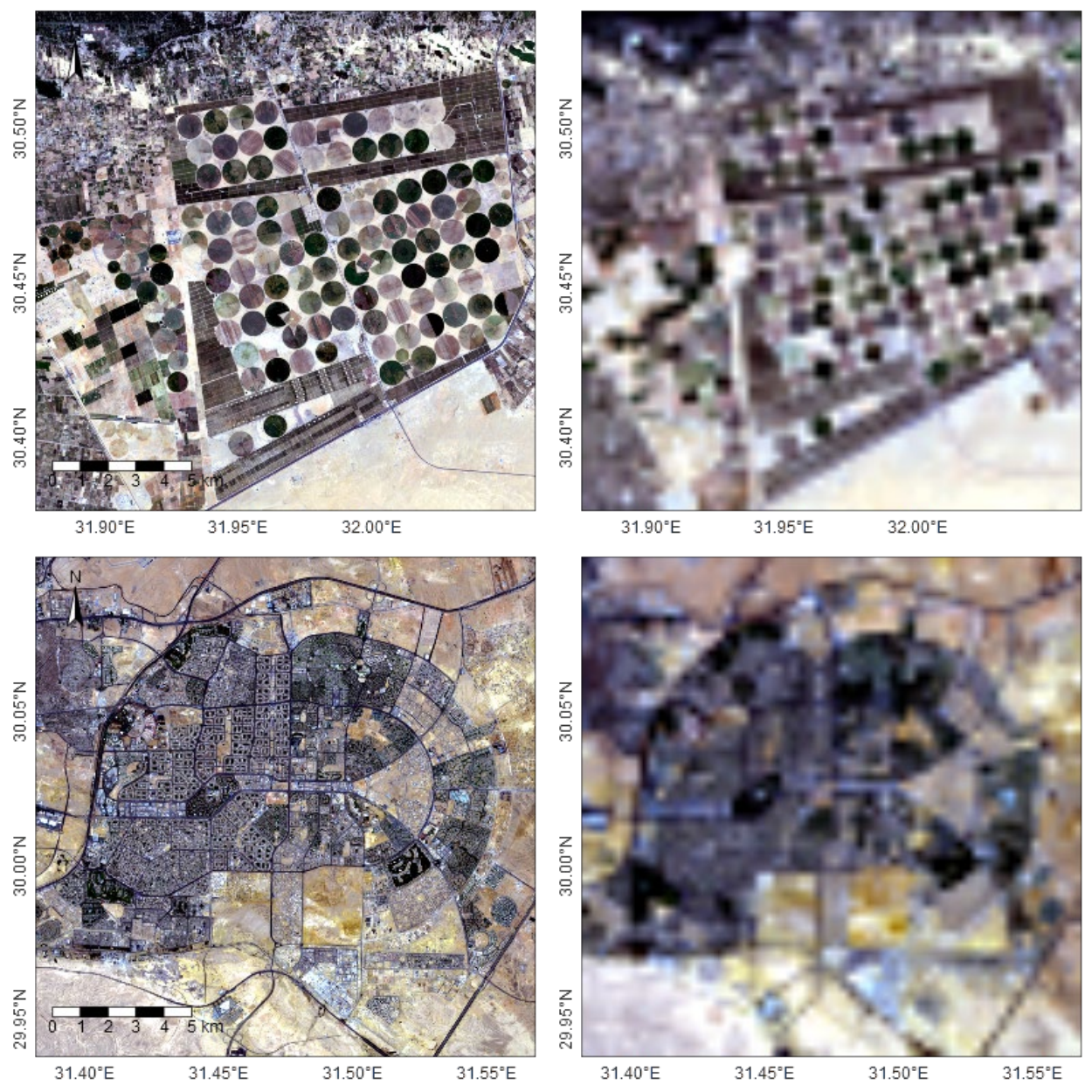}
    \caption{RGB representation of the fine-scale (left) and aggregated (right) imagery for the Croplands (top) and New Cairo (bottom) datasets. Images correspond to pixel-by-pixel averages over the time-series.}
    \label{fig:roisML}
\end{figure}

% Overall description
The selected regions correspond to 18 km by 18 km sites clipped from the Sentinel-2 Multispectral Imager (MSI) \citep{s2msi} and Sentinel-3 Synergy (SYN) \citep{s3syn} time-series products intersecting the Sentinel-2 MSI T36RUU tile. The period of analysis spans from September $25^{th}$, $2022$ to August $6^{th}$, $2023$ in both sites, with a total number of $15$ and $19$ images for the Croplands and New Cairo regions, respectively (see Table \ref{table:tseries}).

\renewcommand{\arraystretch}{0.8}
\begin{table}[!h]
    \caption{Time-series of images in both regions.}
    \begin{center}
    %\resizebox{!}{0.28\textheight}{
            \begin{tabular}{l|c|c}
                \hline
                \textbf{No.} & \textbf{Croplands} & \textbf{New Cairo} \\
                \hline
                1  & 2022-09-25  & 2022-09-25  \\
                2  & 2022-09-30  & 2022-09-30  \\
                3  & 2022-11-19  & 2022-10-15  \\
                4  & 2022-12-29  & 2022-11-19  \\
                5  & 2023-01-18  & 2022-12-29  \\
                6  & 2023-04-03  & 2023-01-18  \\
                7  & 2023-06-02  & 2023-03-04  \\
                8  & 2023-06-12  & 2023-04-03  \\
                9  & 2023-06-17  & 2023-04-18  \\
                10 & 2023-06-22  & 2023-06-02  \\
                11 & 2023-06-27  & 2023-06-12  \\
                12 & 2023-07-02  & 2023-06-17  \\
                13 & 2023-07-17  & 2023-06-22  \\
                14 & 2023-07-22  & 2023-06-27  \\
                15 & 2023-08-01  & 2023-07-02  \\
                16 & 2023-08-06  & 2023-07-12  \\
                17 & -           & 2023-07-17  \\
                18 & -           & 2023-07-22  \\
                19 & -           & 2023-08-01  \\
                20 & -           & 2023-08-06  \\
                \hline
        \end{tabular}%}
    \end{center}
    \label{table:tseries}
\end{table}

% Resolutions
Fine images for both regions are of $900 \times 900$ pixels, with a spatial resolution of 20 m defined on the original Sentinel-2 MSI grid. These images are then aggregated into coarse-resolution images by averaging groups of $15 \times 15$ fine-resolution pixels into 300 m pixels, corresponding to $F_\mathbf{T}(\mathbf{S}_F,\mathbf{b})$ in Section \ref{sec:ABSIS}. Sentinel-3 SYN data are provided in a non-regularly gridded format, and to ensure comparability, we transform them into a regular grid by projecting the data into the default Sentinel-3 SYN grid. Those images are then filled and smoothed using the Interpolation of Mean Anomalies (IMA) method \citep{militino2019ima} to produce images corresponding to $C_{\mathbf{T}}(\mathbf{S},\mathbf{b})$ in Section \ref{sec:std_methods}. Table \ref{tab:wavelengths} summarizes the bands used for the analysis.

\begin{table}[!ht]
		\caption{Band wavelengths of Sentinel-2 MSI and Sentinel-3 SYN products.}
		\begin{center}
			%\resizebox{\textwidth}{!}{
				\begin{tabular}{ l|c|c|c|c }
					\Xhline{4\arrayrulewidth}
					\multirow{2}{*}{\textbf{Name}} & \multicolumn{2}{c|}{\textbf{Sentinel-2 (MSI)}} & \multicolumn{2}{c}{\textbf{Sentinel-3 (SYN)}}\\
					\cline{2-5}
					\multirow{2}{*}{} & Wavelength & Channel Code &  Wavelength & Channel Code\\
					\hline
					Red                          & $650-680 \ nm$   & \texttt{B04} & $660-670 \ nm$   & \texttt{SDR\_Oa08}\\
					Green                        & $543-578 \ nm$   & \texttt{B03}  & $555-565 \ nm$   & \texttt{SDR\_Oa06}\\
					Blue                         & $458-523 \ nm$   & \texttt{B02}   & $485-495 \ nm$   & \texttt{SDR\_Oa04}\\
					NIR          & $855-875 \ nm$   & \texttt{B8A}   & $855-875 \ nm$   & \texttt{SDR\_Oa17}\\
					\Xhline{4\arrayrulewidth}
			\end{tabular}%}
		\end{center}
		\label{tab:wavelengths}
	\end{table}

\subsection{Standardization experiments} \label{sec:stdassessment}

Evaluating the effectiveness of the pairwise upscaling is straightforward as the similarities correspond to the optimal correlation coefficients obtained for the search algorithm. To assess the generalized upscaling, we evaluate the similarity using the parameters obtained when the image is part of the test set for the cross-validation. We use the similarity between the upscaled and Sentinel-3 images to evaluate the upscaling methods.

Evaluation of the effectiveness of the ABSIS method requires an experimental framework that tests its ability to standardize satellite images across different temporal instances. To achieve this, we conducted a series of experiments in which images at specific target dates were treated as missing and reconstructed using a six-image baseline. We evaluated the ABSIS method by measuring the similarity between the upscaled fine-resolution and the standardized images.

%Experiments
We used a 6-image baseline ($|\mathbf{T}|$ = 6) comprised of the three previous and three posterior images in each experiment for ABSIS. Thus, the first experiment in the Croplands region consists of standardizing the $4^{th}$ image, corresponding to December $29^{th}$, 2022, using the 3 previous and 3 posterior images as baseline, that is, $\mathbf{T}$ comprises dates No. $1$ to $3$ and $5$ to $7$ in Table \ref{table:tseries}, respectively. The evaluation metric chosen to assess similarity is the linear correlation, measured by the Pearson's correlation coefficient.

\begin{table} [ht]
\centering
\caption{Similarities for the Croplands region using the proposed standardization methods. Each column contains the correlation coefficient for a diferent method, 1v1 ups and gral ups represent the pairwise and generalized upscaling, respectively, while the last column corresponds to the ABSIS method.}
\label{tab:std_ccs_crp}
%\resizebox{0.6\textwidth}{0.2\textheight}{
\begin{tabular}{lrrr}
  \hline
Date & 1v1 ups $\rho$ & gral ups $\rho$ & ABSIS $\rho$ \\ 
    \hline
    2022-09-25 & \textbf{0.9296} & 0.9283 &  \\ 
    2022-09-30 & \textbf{0.9315} & 0.9305 &  \\ 
    2022-11-19 & \textbf{0.9599} & 0.9593 &  \\ 
    2022-12-29 & \textbf{0.9539} & 0.9508 & 0.8428 \\ 
    2023-01-18 & \textbf{0.9300} & 0.9293 & 0.8198 \\ 
    2023-04-03 & \textbf{0.9416} & 0.9408 & 0.6979 \\ 
    2023-06-02 & \textbf{0.9352} & 0.9168 & 0.7697 \\ 
    2023-06-12 & \textbf{0.9592} & 0.9514 & 0.8587 \\ 
    2023-06-17 & \textbf{0.9591} & 0.9584 & 0.9070 \\ 
    2023-06-22 & \textbf{0.9406} & 0.9397 & 0.9334 \\ 
    2023-06-27 & \textbf{0.9562} & 0.9540 & 0.9450 \\ 
    2023-07-02 & \textbf{0.9570} & 0.9553 & 0.9286 \\ 
    2023-07-17 & \textbf{0.9272} & 0.9247 & 0.9096 \\ 
    2023-07-22 & \textbf{0.9340} & 0.9317 &  \\ 
    2023-08-01 & \textbf{0.9578} & 0.9564 &  \\ 
    2023-08-06 & \textbf{0.9400} & 0.9351 &  \\ 
   \hline
\end{tabular}%}
\end{table}

% Standardization results
Similarities obtained by the different methods for the Croplands and New Cairo regions are summarized in Tables \ref{tab:std_ccs_crp} and \ref{tab:std_ccs_nc}, respectively. Both upscaling methods yield linear correlation coefficients greater than $0.91$ for all dates in the Croplands region, in contrast to the ABSIS method, which has a minimum correlation of $0.7$ and exhibits a smaller similarity for all the images. Both upscaling methods provide similar correlations for the New Cairo region. However, the ABSIS method performs much better in this region compared to the Croplands region, providing a minimum similarity of $0.86$, close to that for the upscaling, and even beating the upscaling methods in some instances; for example, the ABSIS standardized image has a correlation of $0.98$ for July $2^{nd}$, 2023. Note that there is no ABSIS sharpened image for the three first and three last dates of the time-series, since those are required to build the corrected images using ABSIS.

\begin{table} [!h]
\centering
\caption{Similarities for the New Cairo region using the proposed standardization methods. Each column contains the correlation coefficient for a diferent method, 1v1 ups and gral ups represent the pairwise and generalized upscaling, respectively, while the last column corresponds to the ABSIS method.}
\label{tab:std_ccs_nc}
%\resizebox{0.6\textwidth}{0.2\textheight}{
\begin{tabular}{lrrr}
  \hline
Date & 1v1 ups $\rho$ & gral ups $\rho$ & ABSIS $\rho$ \\ 
  \hline
  2022-09-25 & \textbf{0.9630} & 0.9629 & - \\ 
  2022-09-30 & \textbf{0.9347} & 0.9335 & - \\ 
  2022-10-15 & \textbf{0.9121} & 0.9108 & - \\ 
  2022-11-19 & \textbf{0.9467} & 0.9464 & 0.9097 \\ 
  2022-12-29 & \textbf{0.9412} & 0.9404 & 0.9401 \\ 
  2023-01-18 & 0.9122 & 0.9106 & \textbf{0.9147} \\ 
  2023-03-04 & \textbf{0.9301} & 0.9297 & 0.8612 \\ 
  2023-04-03 & 0.8708 & 0.8704 & \textbf{0.8821} \\ 
  2023-04-18 & \textbf{0.9343} & 0.9336 & 0.9012 \\ 
  2023-06-02 & 0.9388 & 0.9276 & \textbf{0.9400} \\ 
  2023-06-12 & 0.9324 & 0.9270 & \textbf{0.9509} \\ 
  2023-06-17 & 0.9345 & 0.9326 & \textbf{0.9737} \\ 
  2023-06-22 & 0.9293 & 0.9273 & \textbf{0.9698} \\ 
  2023-06-27 & 0.9357 & 0.9334 & \textbf{0.9715} \\ 
  2023-07-02 & 0.9091 & 0.9060 & \textbf{0.9815} \\ 
  2023-07-12 & 0.9209 & 0.9184 & \textbf{0.9761} \\ 
  2023-07-17 & 0.9061 & 0.9016 & \textbf{0.9622} \\ 
  2023-07-22 & \textbf{0.8892} & 0.8873 & - \\ 
  2023-08-01 & \textbf{0.9474} & 0.9471 & - \\ 
  2023-08-06 & \textbf{0.9251} & 0.9224 & - \\ 
   \hline
\end{tabular}%}
\end{table}

\subsection{Fusion experiments} \label{sec:ustfipassessment}

%% TODO - More detail about upscaling process
To assess the effect of the different standardization methods we apply USTFIP using images from the three previous and posterior dates, as in Section \ref{sec:stdassessment}, introducing some changes to the Coarse Harmonization (CH) step (see Figure \ref{fig:ustfip_flowchart}), depending on the standardization approach. For the upscaling approaches, we modify phase (1) Upscaling, which originally involved direct bilinear interpolation, with the upscaling methods proposed in Sections \ref{sec:1v1ups} and \ref{sec:gralups}. In the case of the ABSIS method, we use it to replace the whole CH step. Phase (1) Upscaling is substituted by an average aggregation, and ABSIS gives the radiometrically corrected image resulting from phase (2). The evaluation metrics chosen for the experiments are the RMSE to evaluate the spectral accuracy and the Robert's Edge (Edge) to evaluate the spatial accuracy. Both metrics are then averaged across the bands described in Table \ref{tab:wavelengths}.

\begin{table} [!t]
    \centering
    \caption{Spectral accuracy metrics for the Croplands region, each column contains the fusion RMSE obtained using a different standardization method.}
    \begin{tabular}{lrrrr}
      \hline
    Date & CH & 1v1 ups & gral ups & ABSIS \\ 
    \hline
    2022-12-29 & 0.0603 & 0.0591 & 0.0609 & \textbf{0.0602} \\ 
    2023-01-18 & 0.0647 & 0.0642 & 0.0622 & \textbf{0.0575} \\ 
    2023-04-03 & 0.0702 & 0.0734 & \textbf{0.0754} & 0.0791 \\ 
    2023-06-02 & 0.0777 & 0.0724 & \textbf{0.0448} & 0.0681 \\ 
    2023-06-12 & 0.0567 & 0.0550 & \textbf{0.0409} & 0.0496 \\ 
    2023-06-17 & 0.0482 & 0.0425 & \textbf{0.0336} & 0.0420 \\ 
    2023-06-22 & 0.0580 & 0.0480 & 0.0506 & \textbf{0.0317} \\ 
    2023-06-27 & 0.0655 & 0.0552 & 0.0558 & \textbf{0.0356} \\ 
    2023-07-02 & 0.0637 & 0.0556 & 0.0543 & \textbf{0.0363} \\ 
    2023-07-17 & 0.1266 & 0.1054 & 0.1009 & \textbf{0.0396} \\ 
   \hline
   Mean & 0.0692 & 0.0631 & 0.0579 & \textbf{0.0500} \\ 
   \hline
    \end{tabular}
    \label{tab:rmse_croplands}
\end{table}

\begin{table} [!t]
    \centering
    \caption{Spatial accuracy metrics for the Croplands region, each column contains the Edge given by a different standardization method and the mean of their absolute values.}
    \begin{tabular}{lrrrr}
        \hline
         Date & CH & 1v1 ups & gral ups & ABSIS \\ 
  \hline
    2022-12-29 & -0.3992 & -0.2289 & -0.4616 & \textbf{-0.0953} \\ 
    2023-01-18 & -0.3741 & -0.2242 & -0.4000 & \textbf{-0.1177} \\ 
    2023-04-03 & -0.5191 & -0.3801 & -0.5710 & \textbf{-0.3048} \\ 
    2023-06-02 & -0.3461 & -0.2220 & -0.2854 & \textbf{0.0133} \\ 
    2023-06-12 & -0.3833 & -0.1928 & -0.4336 & \textbf{-0.0772} \\ 
    2023-06-17 & -0.2830 & -0.0926 & -0.2627 & \textbf{0.0874} \\ 
    2023-06-22 & -0.5910 & -0.3689 & -0.4620 & \textbf{-0.1216} \\ 
    2023-06-27 & -0.6024 & -0.4197 & -0.4806 & \textbf{-0.1433} \\ 
    2023-07-02 & -0.6042 & -0.4521 & -0.4934 & \textbf{-0.1554} \\ 
    2023-07-17 & -0.4921 & -0.3772 & -0.3782 & \textbf{-0.0889} \\
   \hline
   Mean($|\cdot|$) & 0.4595 & 0.2958 & 0.4228 & \textbf{0.1205} \\ 
   \hline
    \end{tabular}
    \label{tab:edge_croplands}
\end{table}

% Croplands fusion
Tables \ref{tab:rmse_croplands} and \ref{tab:edge_croplands} show the spectral and spatial accuracy metrics obtained for the experiments in the Croplands region for the different standardization methods described in Section \ref{sec:std_methods}. Each table also contains the results obtained using the original Coarse harmonization step as a baseline for the assessment and the mean accuracy for easier comparison. The pairwise upscaling approach decreases RMSE by $8.82\%$ relative to the original CH step, while the generalized upscaling approach yields a decrease of $16.33\%$. However, greater improvement is observed in the spatial accuracies, reducing the mean difference on the Robert's Edge feature by $35.63\%$ for the pairwise upscaling, while the general upscaling exhibits a decrease of $7.99\%$. Using ABSIS to standardize the images yields an acceptable increase in the spectral accuracy compared to the upscaling methods, with a decrease of $27.75\%$ in the average RMSE relative to the original Coarse Harmonization (CH) step, and a great increase regarding the spatial accuracy, decreasing the mean difference in the Robert's Edge Feature by $73.76\%$.

\begin{table} [!t]
    \centering
    \caption{Spectral accuracy metrics for the New Cairo region, each column contains the RMSE given by a different standardization method.}
    \begin{tabular}{lrrrr}
  \hline
Date & CH & 1v1 ups & gral ups & ABSIS \\ 
  \hline
    2022-11-19 & 0.0477 & 0.0385 & 0.0400 & \textbf{0.0316} \\ 
    2022-12-29 & 0.0502 & 0.0422 & 0.0433 & \textbf{0.0346} \\ 
    2023-01-18 & 0.0543 & 0.0479 & 0.0544 & \textbf{0.0391} \\ 
    2023-03-04 & 0.1022 & 0.1065 & 0.0823 & \textbf{0.0398} \\ 
    2023-04-03 & 0.0520 & 0.0381 & 0.0426 & \textbf{0.0387} \\ 
    2023-04-18 & 0.0438 & 0.0297 & 0.0353 & \textbf{0.0321} \\ 
    2023-06-02 & 0.0429 & 0.0333 & 0.0400 & \textbf{0.0285} \\ 
    2023-06-12 & 0.0330 & 0.0212 & 0.0265 & \textbf{0.0233} \\ 
    2023-06-17 & 0.0275 & 0.0208 & \textbf{0.0208} & 0.0247 \\ 
    2023-06-22 & 0.0464 & 0.0370 & 0.0363 & \textbf{0.0204} \\ 
    2023-06-27 & 0.0507 & 0.0379 & 0.0446 & \textbf{0.0216} \\ 
    2023-07-02 & 0.0505 & 0.0398 & 0.0438 & \textbf{0.0199} \\ 
    2023-07-12 & 0.0463 & 0.0355 & 0.0395 & \textbf{0.0196} \\ 
    2023-07-17 & 0.1312 & 0.1147 & 0.0893 & \textbf{0.0202} \\ 
   \hline
   Mean & 0.0556 & 0.0459 & 0.0456 & \textbf{0.0281} \\
   \hline
\end{tabular}
    \label{tab:rmse_newcairo}
\end{table}

\begin{table} [!h]
    \centering
    \caption{Spatial accuracy metrics for the New Cairo region, each column contains the Edge given by a different standardization method and the mean of their absolute values.}
    \begin{tabular}{lrrrr}
      \hline
    Date & CH & 1v1 ups & gral ups & ABSIS \\ 
    \hline
    2022-11-19 & -0.5396 & -0.3412 & -0.3608 & \textbf{-0.1333} \\ 
    2022-12-29 & -0.5087 & -0.3021 & -0.3353 & \textbf{-0.1012} \\ 
    2023-01-18 & -0.5169 & -0.3481 & -0.3518 & \textbf{-0.1214} \\ 
    2023-03-04 & -0.2326 & 0.0147 & -0.2021 & \textbf{-0.0832} \\ 
    2023-04-03 & -0.5731 & -0.3259 & -0.4364 & \textbf{-0.1946} \\ 
    2023-04-18 & -0.5215 & -0.2204 & -0.3675 & \textbf{-0.0973} \\ 
    2023-06-02 & -0.5732 & -0.3484 & -0.4421 & \textbf{-0.1302} \\ 
    2023-06-12 & -0.4250 & -0.0724 & -0.2552 & \textbf{-0.0090} \\ 
    2023-06-17 & -0.3237 & -0.0007 & -0.1744 & \textbf{0.1178} \\ 
    2023-06-22 & -0.3326 & -0.1728 & -0.2625 & \textbf{-0.0882} \\ 
    2023-06-27 & -0.5972 & -0.3686 & -0.4815 & \textbf{-0.1214} \\ 
    2023-07-02 & -0.6029 & -0.4113 & -0.4810 & \textbf{-0.1200} \\ 
    2023-07-12 & -0.5582 & -0.3474 & -0.4373 & \textbf{-0.0624} \\ 
    2023-07-17 & -0.3898 & -0.2974 & -0.2975 & \textbf{-0.0667} \\
   \hline
       Mean($|\cdot|$) & 0.4782 & 0.2551 & 0.3489 & \textbf{0.1033} \\
    \hline
    \end{tabular}
    \label{tab:edge_newcairo}
\end{table}

Tables \ref{tab:rmse_newcairo} and \ref{tab:edge_newcairo} give the accuracies for the experiments in the New Cairo region. Pairwise upscaling yields a $17.45\%$ decrease in RMSE, while the generalized upscaling provides slightly increased accuracy on average, with an increase of $17.99\%$ regarding spectral accuracy, both relative to the original CH step. The upscaling methods exhibit slightly different behaviors in this region, decreasing the mean of the Edge metric by $46.65\%$ and $14.83\%$ for the pairwise and generalized upscaling, respectively, compared to the $27.04\%$ and $24.82\%$ found for the Croplands region. Feeding the ABSIS standardized images, in addition to the resampled Sentinel-2 images, into the fusion model greatly increases the average spectral and spatial accuracies, yielding decreases of $49.46\%$ and $78.40\%$ in the metrics, respectively.

\section{Discussion} \label{sec:discussion}

% Summary
Multispectral satellite images from different sensors usually exhibit limited linear correlation, which prevents the use of STIF methods to capture linear relationships between images captured on the same dates. Consequently, additional corrections are essential to enhance this correlation and ease the generation of sets of images with matching spatial and spectral resolutions captured by different sensors. We propose two different approaches to solve the standardization problem. First, an upscaling approach that degrades fine resolution images to resemble the coarse resolution ones while accounting for the different challenges in multi-sensor fusion. Second, an anomaly-based approach that seeks to sharpen the coarse resolution images so they more closely resemble the aggregated fine-resolution images.

While this research focuses on STIF, the proposed methods address common challenges in many multi-source remote sensing tasks, such as geometric misalignment, spectral differences between sensors and spatial resolution mismatches. By explicitly targeting these issues, the proposed methodologies offer a general and robust foundation for diverse multi-sensor scenarios.

The alignment and spectral compatibility problems may yield non-linear relationships between observed data. This problem is usually solved by using more complex models \citep{vivone2025DLreview}, with deep learning-based approaches emerging as the most prominent example \citep{MOOSAVI2015NN, SHAO2019DLfusion, CAI2022dNN}. Another promising direction is spatiotemporal-spectral fusion, which also integrates the spectral characteristics of the data \citep{Shen2012STSf}. \cite{Jiang2022STSf} proposed a framework that combined both approaches. Meanwhile, \cite{MENG2025GLOSTFM} explicitly addressed the multi-sensor discrepancies by evaluating them at each level of a Laplacian pyramid to improve the overall reconstruction. However, more complex models often come with increased computational costs and reduced interpretability. In contrast, theoretically more straightforward fusion methods, which are more explainable and make more efficient use of data, often face practical limitations due to inadequate preprocessing. Through the use of well-implemented standardization techniques, even basic fusion algorithms can achieve significantly improved results while maintaining low computational overhead and high interpretability.

\begin{figure} [!h]
    \centering
    \includegraphics[width=\linewidth]{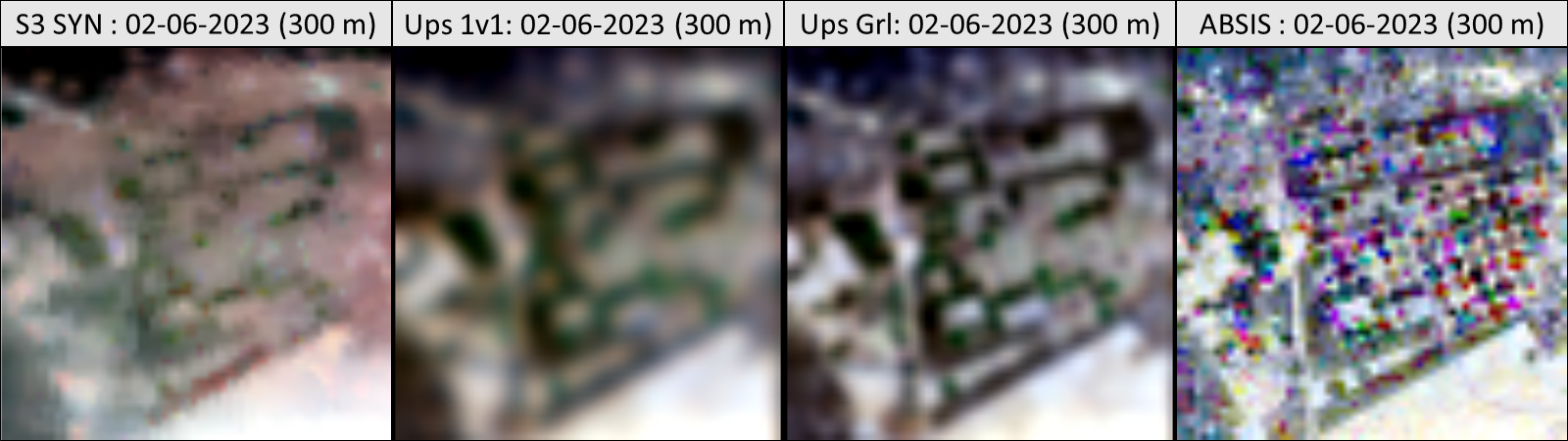}
    \caption{RGB representations of the standardization results for the Croplands region. From left to right, the smoothed Sentinel-3 image, the pairwise upscaled image, the generalized upscaling image and the ABSIS standardized image.}
    \label{fig:ups_crp}
\end{figure}

\begin{figure} [!h]
    \centering
    \includegraphics[width=\linewidth]{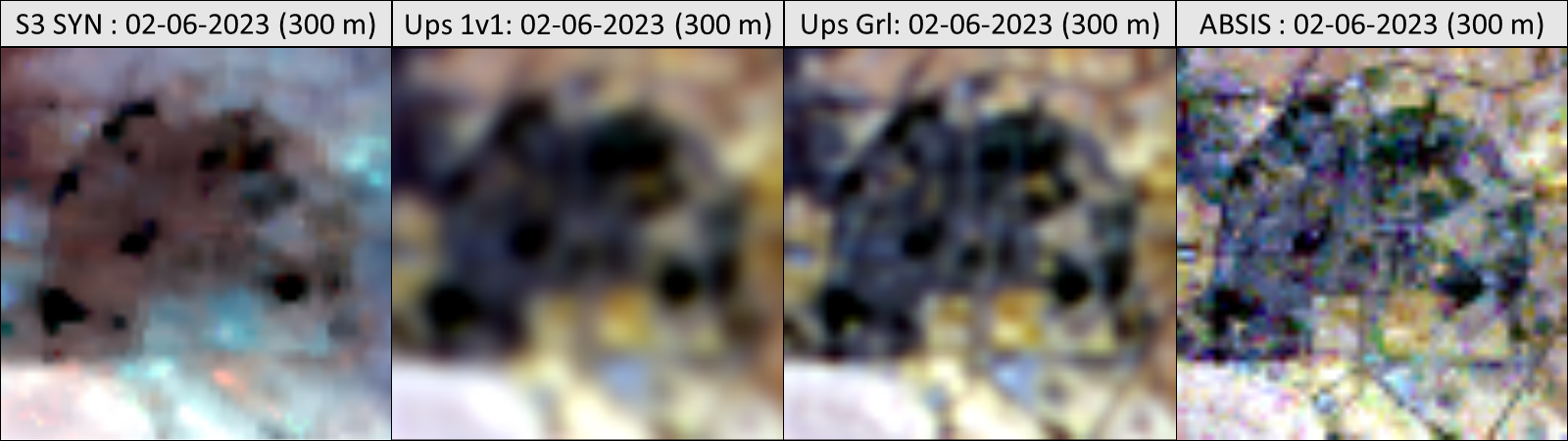}
    \caption{RGB representations of the standardization results for the New Cairo region. From left to right, the smoothed Sentinel-3 image, the pairwise upscaled image, the generalized upscaling image and the ABSIS standardized image.}
    \label{fig:ups_nc}
\end{figure}

% Upscaling discussion
As shown in Section \ref{sec:stdassessment} the upscaling methods are superior in terms of increasing correlation, as shown by the fact that the linear correlations are larger on average in Tables \ref{tab:std_ccs_crp} and \ref{tab:std_ccs_nc}, as a result of explicitly solving the multi-sensor fusion challenges, such as the geometric and capture discrepancies. In addition, the upscaling methods are more robust to region particularities, as opposed to the great differences found in the ABSIS results between the Croplands and New Cairo region. Those differences probably arise from the aliasing effect, which renders the aggregated fine-resolution images unable to represent the circle-like shapes present in the fine-resolution images, which were incorrectly represented as rectangles in the aggregated fine-resolution images (see Figure \ref{fig:roisML}). This makes the aggregated images much less similar to the coarse resolution ones on a per-pixel basis, significantly hindering the strategy of combining the spatial pattern with the particular anomaly, as shown by the speckle-like distortion from the ABSIS image in Figure \ref{fig:ups_crp}, not seen in Figure \ref{fig:ups_nc}. It is also worth noting that even if optimizing the pairwise correlation yields larger linear correlations, it does not account for the target image quality. Thus, in case of low quality images, such as the Sentinel-3 SYN images from Figures \ref{fig:ups_crp} and \ref{fig:ups_nc}, the optimal pairwise upscaling may induce oversmoothing to increase the linear correlation, as shown by the differences between the images upscaled through pairwise and generalized upscaling in Figures \ref{fig:ups_crp} and \ref{fig:ups_nc}.

\begin{figure} [!h]
    \centering
    \includegraphics[width=\linewidth]{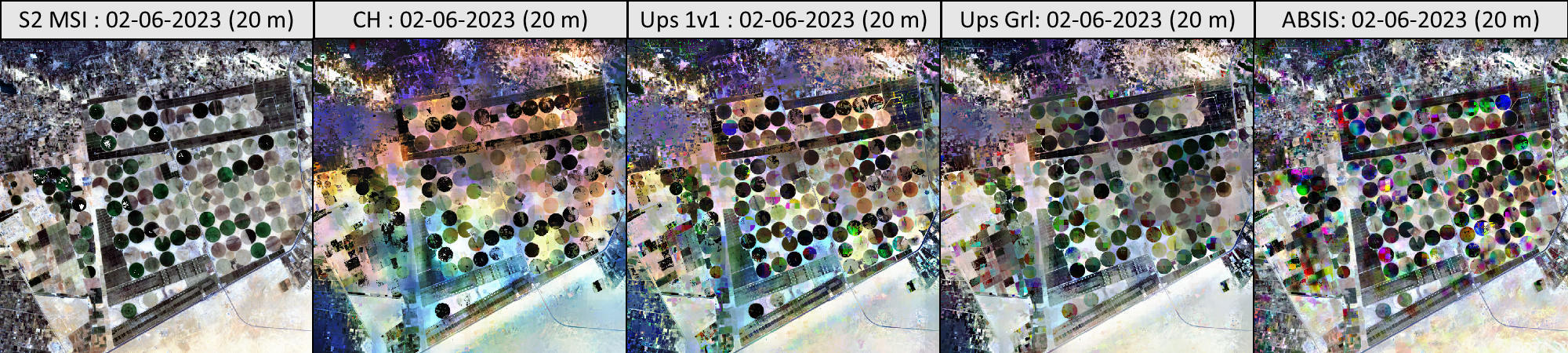}
    \caption{RGB representations of the USTFIP results for the Croplands region. From left to right, the original Sentinel-2 image, the base fused image, using the original CH step, the fused image using pairwise upscaling, the fusion result using the generalized upscaling and the result of using the ABSIS standardized image.}
    \label{fig:absis_crp}
\end{figure}

\begin{figure} [!h]
    \centering
    \includegraphics[width=\linewidth]{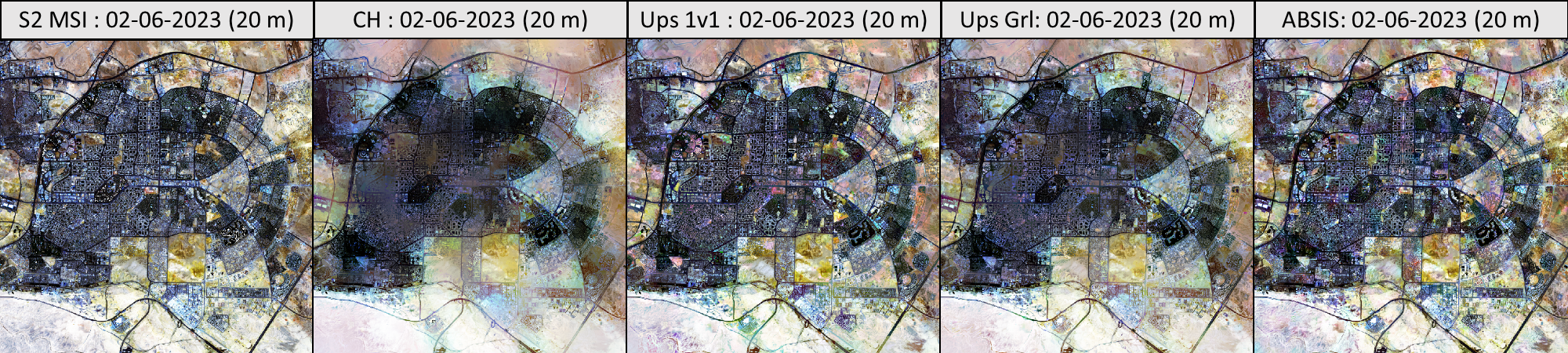}
    \caption{RGB representations of the USTFIP results for the New Cairo region. From left to right, the original Sentinel-2 image, the base fused image, using the original CH step, the fused image using pairwise upscaling, the fusion result using the generalized upscaling and the result of using the ABSIS standardized image.}
    \label{fig:absis_nc}
\end{figure}

% Fusion discussion
As discussed previously in this section, the upscaling methods seem to be better suited to increase the linear correlation. However, as shown in Section \ref{sec:ustfipassessment}, feeding USTFIP with the ABSIS-corrected image in addition to aggregated fine resolution images yielded more accurate results in terms of spectral accuracy, as shown by Tables \ref{tab:rmse_croplands} and \ref{tab:rmse_newcairo}. This improvement is even greater when considering spatial accuracy, generating images whose spatial patterns resemble much more closely those of the target coarse resolution images, probably due to the fact that differences measured from aggregated images then transfer more readily to the fine-resolution, as shown by the fact that the shapes in the fused images obtained from the ABSIS-corrected images are much sharper, and thus, closer to the original Sentinel-2 ones. This behavior is expected since we are using the aggregated fine-resolution as the coarse-resolution baseline. Thus, the only difference between the coarse- and fine- baselines is the averaging, making the transference of the predicted temporal change straightforward. Slight differences in the band definitions from each sensor may also have a negative effect on the fusion, even in the case of bands with the same wavelength at both resolutions, such as the NIR band from our experiments (see Table \ref{tab:wavelengths}). Since the spectral responses captured by both sensors are different, the result of digitizing the same segment of the spectrum is different (i.e., the same wavelength may be representing slightly different responses). This may induce nonlinearity in the relationship between the coarse- and fine- resolution images, hindering information sharing between resolutions. The PSF simulations reduce this effect to an extent. However, solving this problem from the upscaling point of view would require great changes to introduce some kind of multi-band (or even multi-source) approach along with the co-registration and PSF simulation. In contrast, ABSIS solves the issue implicitly by borrowing information directly from the fine-resolution images, which results in images that retain the spectral behavior to a greater extent.

% General discussion
Each standardization approach has its own advantages and pitfalls. Upscaling requires strategies to find the optimal parameters in a timely manner, while sharpening the coarse resolution images requires a time-series of paired images. The proposed upscaling methods tackle explicitly the capture geometry differences and the misalignment, yielding images that are more similar in terms of the linear correlation. Meanwhile, the main advantage of ABSIS relies on the fact that the input coarse-resolution images maintain some fine-resolution information, making the transfer of the temporal changes more straightforward. In addition, since ABSIS borrows strength from the time-series of fine resolution images, the spectral behavior is closer to those images, reducing the influence of the sensor-induced spectral differences into the final prediction, while also implicitly reducing the effect of geometric discrepancies.

\section{Conclusions} \label{sec:conclusion}

Multispectral satellite images from different sensors usually exhibit limited linear correlation, which prevents the use of STIF methods to capture linear relationships between images captured on the same dates. Consequently, additional actions are essential to enhance this correlation and ease the generation of sets of images with matching spatial and spectral resolutions captured by different sensors. The long revisit times of fine-resolution sensors make their images scarce, which makes solving this problem more challenging since vast numbers of data are often required for different types of feature extraction, either in terms of pseudo-invariant areas that improve radiometric normalization \citep{pons2014automatic} or as part of a deep-learning framework. Although such approaches provide good results in cases involving large-scale predictions and large numbers of data, they may not be appropriate for STIF methods based on transferring temporal changes found in coarse-resolution images to predict fine-resolution ones, such as in \cite{fitfc2018} and \cite{goyena2023ustfip}. 

Our research explored two different approaches to tackle the standardization problem. First, we explored an upscaling approach that tackles explicitly most of the issues found in multi-sensor fusion, such as spatial discrepancies and capture geometry differences. This approach seeks to degrade the fine resolution images so that they resemble the coarse resolution images, providing a series of coarse resolution images that is more similar overall. Second, we explored decomposing the images to blend the overall features found in the fine-resolution image series with the distinctive properties of a particular coarse-resolution image. This approach, termed Anomaly-Based Satellite Image Standardization (ABSIS) aims to produce images that resemble the outcome of resampling the fine-resolution images. We compared a sharpening based standardization ABSIS method to different upscaling methods aiming to downgrade the fine resolution image using reduced number of data.

Our experiments demonstrate that the upscaling approach yields image sets that are more linearly similar overall by approximating the coarse resolution images. On the other hand, ABSIS greatly increases fusion accuracy, both in terms of spectral and spatial accuracies, outperforming the upscaling methods, which explicitly tackle the multi-sensor fusion challenges, hinting that separating the anomalies from the overall features eases the relationship transfer from the coarse to the fine resolution, which increases spectral accuracy, reducing the mean RMSE by up to $49.46\%$ compared to the original Coarse Harmonization step. In addition, isolating the pixel-by-pixel pattern makes the spatial correction straightforward, which greatly increases the spatial accuracy, reducing the difference in the Robert's Edge feature up to $78.40\%$.

\newpage

\renewcommand{\arraystretch}{1.2}
\section*{Glossary}
\renewcommand{\arraystretch}{0.9}
\begin{longtable}[h!] {l p{0.8\linewidth}}
    %\centering
    %\begin{tabular}{l p{0.8\linewidth}}
    \hline
    \textbf{Abbreviation} & \textbf{Meaning} \\
    \hline
    ABSIS  & \setstretch{0.5} Anomally-Based Satellite Image Standardization standardization method \\
    CH & \setstretch{0.5} Coarse Harmonization step of the USTFIP method\\
    Edge & \setstretch{0.5} Mean difference of the Robert's Edge feature\\
    ESA & \setstretch{0.5} European Space Agency\\
    Fit-FC & \setstretch{0.5} STIF Method Consisting of regression fitting, residual compensation and spatial filtering\\
    IMA & \setstretch{0.5} Interpolation of the Mean Anomalies gap-filling method\\
    MSI & \setstretch{0.5} Sentinel-2 MultiSpectral Imager mission\\
    NIR & \setstretch{0.5} Near InfraRed band\\
    PSF & \setstretch{0.5} Point Spread Function \\
    RMSE & \setstretch{0.5} Root Mean Squared Error\\
    Sentinel & \setstretch{0.5} Satellite program operated by the ESA\\
    STIF   & \setstretch{0.5} Spatio-Temporal Image Fusion \\
    SYN & \setstretch{0.5} Sentinel-3 SYNergy mission\\
    USTFIP & \setstretch{0.5} Unpaired Spatio-Temporal Fusion of Image Patches - STIF method \\
    \hline
    \textbf{Symbol} & \textbf{Expansion} \\
    \hline
    $C_{\mathbf{T}}(\mathbf{S},\mathbf{b})$ & \setstretch{0.5} Set of coarse-resolution images \\
    $F_{\mathbf{T}}(\mathbf{s},\mathbf{b})$ & \setstretch{0.5} Set of fine-resolution images \\
    $F_{\mathbf{T}}(\mathbf{S}_F,\mathbf{b})$ & \setstretch{0.5} Set of aggregated fine-resolution images \\
    $\mathbf{S}$ & \setstretch{0.5} Set of pixel locations measured at coarse-resolution support \\
    $\mathbf{s}$ & \setstretch{0.5} Set of pixel locations measured at fine-resolution support \\
    $\mathbf{S}_F$ & \setstretch{0.5} Set of pixel locations at the aggregated fine-resolution support \\
    $\mathbf{T}$ & \setstretch{0.5} Set of capture dates \\
    $\mathbf{b}$ & \setstretch{0.5} Set of multispectral bands contained in each image \\
    $g_\sigma$ & \setstretch{0.5} Gaussian filter with standard deviation $\sigma$ \\
    $*$ & \setstretch{0.5} Convolution operation \\
    $\hat{C}_{t_k}$ & \setstretch{0.5} Upscaled image at date $t_k$ \\
    $P_C(\mathbf{S},\mathbf{b})$ & \setstretch{0.5} Coarse-resolution pattern \\
    $P_F(\mathbf{s},\mathbf{b})$ & \setstretch{0.5} Fine-resolution pattern \\
    $a_{t_k}$ & Coarse-resolution anomaly at date $t_k$\\
    $\hat{a}_{t_k}$ & (Aggregated) Fine-resolution anomaly at date $t_k$\\
    $\rho_{t_k}(S_i,b_l)$ & Linear correlation coefficient for a window around pixel $S_i$ for band $b_l$ at date $t_k$ \\
    $\delta(\cdot,\cdot)$ & Selection mask \\
    $\rho(\cdot,\cdot)$ & Pearson's correlation coefficient between images defined at the same support \\
    $\hat{F}^m_{t_0}$ & Fused image obtained using standardized images from method $m$ as input\\
    \hline
    %\end{tabular}
\end{longtable}

\section*{Funding}
	
This research was supported by the Spanish Ministry of Science and Innovation - Spanish Research Agency [PID 2020-113125RB-I00/ MCIN/ AEI/ 10.13039/501100011033 project]; and the Public University of Navarre [Ayudas predoctorales UPNA 2022/2023, Ayudas a la movilidad de doctorado UPNA 2025 and PJUPNA2024-11708 project].

\section*{CRediT Authorship contributions}

Harkaitz Goyena: Validation, Formal Analysis, Investigation, Data curation, Software, Writing -Original draft, Writing - Review and Editing

Peter M Atkinson: Conceptualization, Methodology, Writing - Review and Editing

Unai Pérez-Goya: Conceptualization, Methodology, Writing - Review and Editing

Maria Dolores Ugarte: Conceptualization, Research organization, Writing - Review and Editing, Project administration, Funding acquisition

\section*{Declaration of Competing Interest}

The authors declare that they have no known competing financial interests or personal relationships that could have influenced the work reported in this paper.

\section*{Data and code availability}

Data and code will be available at the following GitHub repository: \url{https://github.com/spatialstatisticsupna/Standardization}.

\section*{Acknowledgments}

This research was completed as part of a visit by Harkaitz Goyena to Lancaster Environment Centre in Lancaster University, under the supervision of Professor Peter M. Atkinson.

\newpage
\appendix

\section{Upscaling algorithm} \label{A1}

Algorithm \ref{alg:greedy} shows the pseudo-code for the greedy hyperparameter used for the one-on-one image upscaling described in Section \ref{sec:1v1ups}.

\begin{algorithm}
    \caption{Pseudo-code for the greedy hyperparameter search} \label{alg:greedy}
    \linespread{1.35}\selectfont
    \scriptsize
    \begin{algorithmic}[1] 
        \State $C_{t_k} \gets$ Sentinel-3 image at date $t_k$ for the band
        \State $F_{t_k} \gets$ Sentinel-2 image at date $t_k$ for the band
        \State opt.pars $\gets (x = 0, y = 0, \sigma = 1)$
        \State shifts $\gets \{(1,0,0),(-1,0,0),(0,1,0),(0,-1,0),(0,0,0.1),(0,0,-0.1)\}$ 
        \State $\hat{C}_{t_k} \gets \text{upscale}(F_{t_k},\text{opt.pars})$
        \State opt.cc $\gets \rho(\hat{C}_{t_k},C_{t_k})$
        \State pre.cc $\gets 0$
 		\While {opt.cc $>$ pre.cc}
            \State pre.cc $\gets$ opt.cc
            \State pre.pars $\gets$ opt.pars
            \For {s in shifts}
                \State pars $\gets$ pre.pars + s
                \State $\hat{C}_{t_k} \gets \text{upscale}(F_{t_k},\text{pars})$
                \State cc $\gets \rho(\hat{C}_{t_k},C_{t_k})$
                \If {cc $>$ opt.cc}
                    \State opt.cc $\gets$ cc
                    \State opt.pars $\gets$ pars
                \EndIf
            \EndFor
        \EndWhile
        \State pre.cc $\gets 0$
        \State opt.off $\gets (x=0, y=0)$
        \While {opt.cc $>$ pre.cc}
            \State pre.cc $\gets$ opt.cc
            \State pre.pars $\gets$ opt.pars
            \For {s in $\{(0.1,0,0),(-0.1,0,0),(0,0.1,0),(0,-0.1,0)\}$}
                \State pars $\gets$ pre.pars + s
                \State $\hat{C}_{t_k} \gets \text{upscale}(F_{t_k},\text{pars})$
                \State cc $\gets \rho(\hat{C}_{t_k},C_{t_k})$
                \If {cc $>$ opt.cc}
                    \State opt.cc $\gets$ cc
                    \State opt.pars $\gets$ pars
                \EndIf
            \EndFor
        \EndWhile
        \State save $\text{upscale}(F_{t_k},\text{opt.pars})$ and opt.pars
	\end{algorithmic}
\end{algorithm}

\newpage
\section{USTFIP flowchart} \label{A2}

\begin{figure} [!h]
    \includegraphics[width=0.9\linewidth]{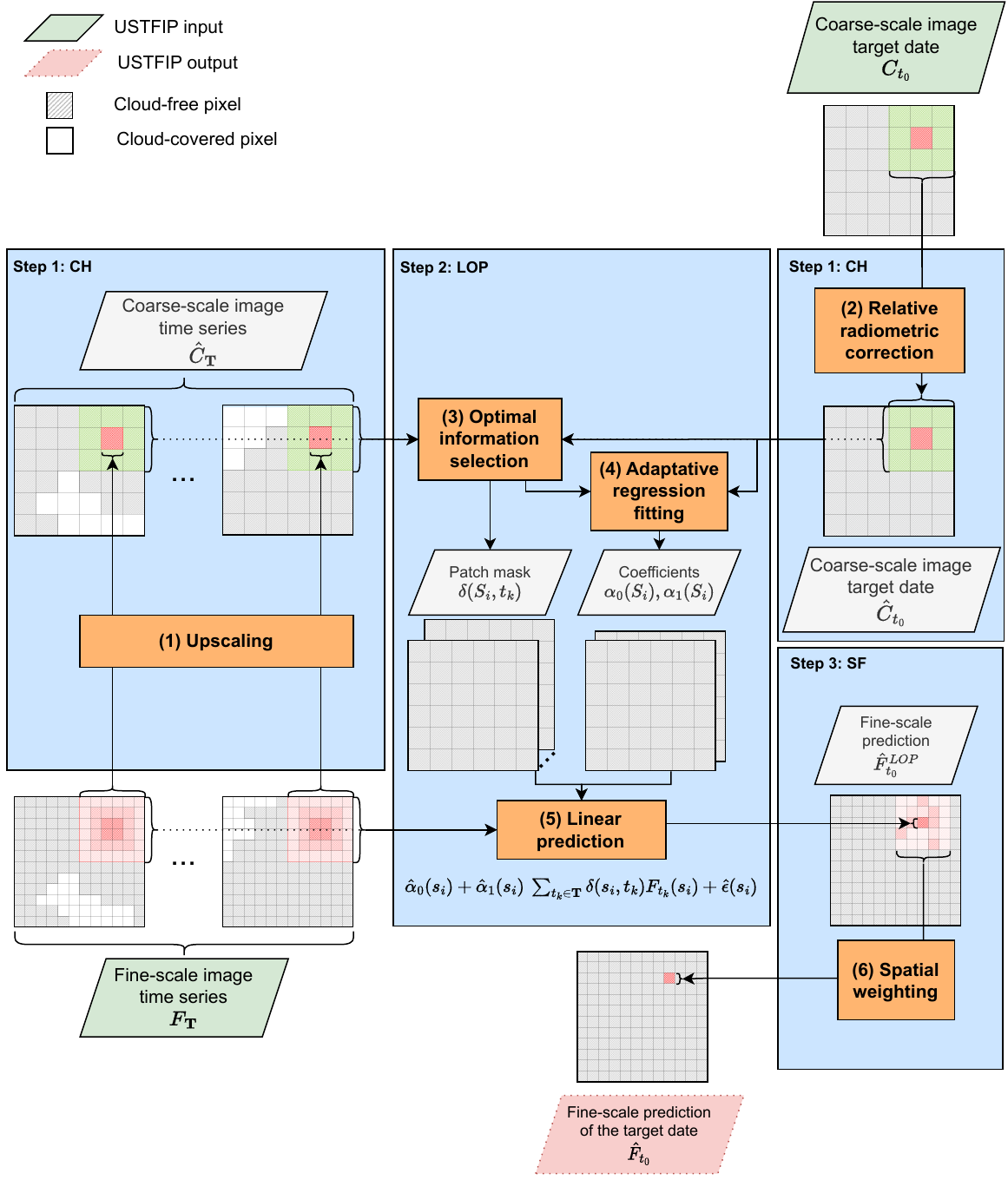}
    \caption{Methodology of the Unpaired Spatio-Temporal Fusion of Image Patches (USTFIP). Step 1 corresponds to Coarse Harmonization (CH), which is modified in this work. Step 2 performs the temporal prediction, while Step 3 spatially smoothens the result.}
    \label{fig:ustfip_flowchart}
\end{figure}

Figure \ref{fig:ustfip_flowchart} summarizes the USTFIP methodology for easier reference.

\newpage
\bibliography{refs}

\begin{thebibliography}{35}
\expandafter\ifx\csname natexlab\endcsname\relax\def\natexlab#1{#1}\fi
\providecommand{\url}[1]{\texttt{#1}}
\providecommand{\href}[2]{#2}
\providecommand{\path}[1]{#1}
\providecommand{\DOIprefix}{doi:}
\providecommand{\ArXivprefix}{arXiv:}
\providecommand{\URLprefix}{URL: }
\providecommand{\Pubmedprefix}{pmid:}
\providecommand{\doi}[1]{\href{http://dx.doi.org/#1}{\path{#1}}}
\providecommand{\Pubmed}[1]{\href{pmid:#1}{\path{#1}}}
\providecommand{\bibinfo}[2]{#2}
\ifx\xfnm\relax \def\xfnm[#1]{\unskip,\space#1}\fi
%Type = Inproceedings
\bibitem[{Abdelrahman et~al.(2011)Abdelrahman, Ali, Elhabian and Farag}]{abdelrahman2011solving}
\bibinfo{author}{Abdelrahman, M.}, \bibinfo{author}{Ali, A.}, \bibinfo{author}{Elhabian, S.}, \bibinfo{author}{Farag, A.A.}, \bibinfo{year}{2011}.
\newblock \bibinfo{title}{Solving geometric co-registration problem of multi-spectral remote sensing imagery using sift-based features toward precise change detection}, in: \bibinfo{booktitle}{Advances in Visual Computing: 7th International Symposium, ISVC 2011, Las Vegas, NV, USA, September 26-28, 2011. Proceedings, Part II 7}, \bibinfo{organization}{Springer}. pp. \bibinfo{pages}{607--616}.
%Type = Article
\bibitem[{Arvidson et~al.(2006)Arvidson, Goward, Gasch and Williams}]{landsat7}
\bibinfo{author}{Arvidson, T.}, \bibinfo{author}{Goward, S.}, \bibinfo{author}{Gasch, J.}, \bibinfo{author}{Williams, D.}, \bibinfo{year}{2006}.
\newblock \bibinfo{title}{Landsat-7 long-term acquisition plan}.
\newblock \bibinfo{journal}{Photogrammetric Engineering \& Remote Sensing} \bibinfo{volume}{72}, \bibinfo{pages}{1137--1146}.
%Type = Article
\bibitem[{Belgiu and Stein(2019)}]{belgiu2019}
\bibinfo{author}{Belgiu, M.}, \bibinfo{author}{Stein, A.}, \bibinfo{year}{2019}.
\newblock \bibinfo{title}{Spatiotemporal image fusion in remote sensing}.
\newblock \bibinfo{journal}{Remote sensing} \bibinfo{volume}{11}, \bibinfo{pages}{818}.
%Type = Article
\bibitem[{Cai et~al.(2022)Cai, Huang and Fung}]{CAI2022dNN}
\bibinfo{author}{Cai, J.}, \bibinfo{author}{Huang, B.}, \bibinfo{author}{Fung, T.}, \bibinfo{year}{2022}.
\newblock \bibinfo{title}{Progressive spatiotemporal image fusion with deep neural networks}.
\newblock \bibinfo{journal}{International Journal of Applied Earth Observation and Geoinformation} \bibinfo{volume}{108}, \bibinfo{pages}{102745}.
\newblock \URLprefix \url{https://www.sciencedirect.com/science/article/pii/S030324342200071X}, \DOIprefix\doi{https://doi.org/10.1016/j.jag.2022.102745}.
%Type = Article
\bibitem[{Chen et~al.(2015)Chen, Huang and Xu}]{chen2015}
\bibinfo{author}{Chen, B.}, \bibinfo{author}{Huang, B.}, \bibinfo{author}{Xu, B.}, \bibinfo{year}{2015}.
\newblock \bibinfo{title}{Comparison of spatiotemporal fusion models: A review}.
\newblock \bibinfo{journal}{Remote Sensing} \bibinfo{volume}{7}, \bibinfo{pages}{1798--1835}.
%Type = Article
\bibitem[{Donaldson and Storeygard(2016)}]{economic2016}
\bibinfo{author}{Donaldson, D.}, \bibinfo{author}{Storeygard, A.}, \bibinfo{year}{2016}.
\newblock \bibinfo{title}{The view from above: Applications of satellite data in economics}.
\newblock \bibinfo{journal}{Journal of Economic Perspectives} \bibinfo{volume}{30}, \bibinfo{pages}{171--98}.
%Type = Article
\bibitem[{Donlon et~al.(2012)Donlon, Berruti, Buongiorno, Ferreira, F{\'e}m{\'e}nias, Frerick, Goryl, Klein, Laur, Mavrocordatos et~al.}]{sentinel3}
\bibinfo{author}{Donlon, C.}, \bibinfo{author}{Berruti, B.}, \bibinfo{author}{Buongiorno, A.}, \bibinfo{author}{Ferreira, M.H.}, \bibinfo{author}{F{\'e}m{\'e}nias, P.}, \bibinfo{author}{Frerick, J.}, \bibinfo{author}{Goryl, P.}, \bibinfo{author}{Klein, U.}, \bibinfo{author}{Laur, H.}, \bibinfo{author}{Mavrocordatos, C.}, et~al., \bibinfo{year}{2012}.
\newblock \bibinfo{title}{The global monitoring for environment and security (gmes) sentinel-3 mission}.
\newblock \bibinfo{journal}{Remote Sensing of Environment} \bibinfo{volume}{120}, \bibinfo{pages}{37--57}.
%Type = Article
\bibitem[{Drusch et~al.(2012)Drusch, Del~Bello, Carlier, Colin, Fernandez, Gascon, Hoersch, Isola, Laberinti, Martimort et~al.}]{sentinel2}
\bibinfo{author}{Drusch, M.}, \bibinfo{author}{Del~Bello, U.}, \bibinfo{author}{Carlier, S.}, \bibinfo{author}{Colin, O.}, \bibinfo{author}{Fernandez, V.}, \bibinfo{author}{Gascon, F.}, \bibinfo{author}{Hoersch, B.}, \bibinfo{author}{Isola, C.}, \bibinfo{author}{Laberinti, P.}, \bibinfo{author}{Martimort, P.}, et~al., \bibinfo{year}{2012}.
\newblock \bibinfo{title}{Sentinel-2: Esa's optical high-resolution mission for gmes operational services}.
\newblock \bibinfo{journal}{Remote sensing of Environment} \bibinfo{volume}{120}, \bibinfo{pages}{25--36}.
%Type = Article
\bibitem[{Emelyanova et~al.(2013)Emelyanova, McVicar, Van~Niel, Li and Van~Dijk}]{emelyanova2013}
\bibinfo{author}{Emelyanova, I.V.}, \bibinfo{author}{McVicar, T.R.}, \bibinfo{author}{Van~Niel, T.G.}, \bibinfo{author}{Li, L.T.}, \bibinfo{author}{Van~Dijk, A.I.}, \bibinfo{year}{2013}.
\newblock \bibinfo{title}{Assessing the accuracy of blending landsat--modis surface reflectances in two landscapes with contrasting spatial and temporal dynamics: A framework for algorithm selection}.
\newblock \bibinfo{journal}{Remote Sensing of Environment} \bibinfo{volume}{133}, \bibinfo{pages}{193--209}.
%Type = Misc
\bibitem[{ESA(2021)}]{s2msi}
\bibinfo{author}{ESA}, \bibinfo{year}{2021}.
\newblock \bibinfo{title}{Copernicus sentinel-2 (processed by esa), 2021, msi level-2a boa reflectance product. collection 1. european space agency.}
\newblock \DOIprefix\doi{10.5270/S2_-znk9xsj}.
%Type = Misc
\bibitem[{ESA(2022)}]{s3syn}
\bibinfo{author}{ESA}, \bibinfo{year}{2022}.
\newblock \bibinfo{title}{European space agency, 2022, level-2 synergy surface directional reflectance, version collection version 2, https://doi.org/10.5270/s3d-643ee90}.
%Type = Article
\bibitem[{Gevaert and Garc{\'\i}a-Haro(2015)}]{coreg2015}
\bibinfo{author}{Gevaert, C.M.}, \bibinfo{author}{Garc{\'\i}a-Haro, F.J.}, \bibinfo{year}{2015}.
\newblock \bibinfo{title}{A comparison of starfm and an unmixing-based algorithm for landsat and modis data fusion}.
\newblock \bibinfo{journal}{Remote sensing of Environment} \bibinfo{volume}{156}, \bibinfo{pages}{34--44}.
%Type = Article
\bibitem[{Ghamisi et~al.(2019)Ghamisi, Rasti, Yokoya, Wang, Hofle, Bruzzone, Bovolo, Chi, Anders, Gloaguen et~al.}]{ghamisi2019}
\bibinfo{author}{Ghamisi, P.}, \bibinfo{author}{Rasti, B.}, \bibinfo{author}{Yokoya, N.}, \bibinfo{author}{Wang, Q.}, \bibinfo{author}{Hofle, B.}, \bibinfo{author}{Bruzzone, L.}, \bibinfo{author}{Bovolo, F.}, \bibinfo{author}{Chi, M.}, \bibinfo{author}{Anders, K.}, \bibinfo{author}{Gloaguen, R.}, et~al., \bibinfo{year}{2019}.
\newblock \bibinfo{title}{Multisource and multitemporal data fusion in remote sensing: A comprehensive review of the state of the art}.
\newblock \bibinfo{journal}{IEEE Geoscience and Remote Sensing Magazine} \bibinfo{volume}{7}, \bibinfo{pages}{6--39}.
%Type = Article
\bibitem[{Goyena et~al.(2023)Goyena, P{\'e}rez-Goya, Montesino-SanMartin, Militino, Wang, Atkinson and Ugarte}]{goyena2023ustfip}
\bibinfo{author}{Goyena, H.}, \bibinfo{author}{P{\'e}rez-Goya, U.}, \bibinfo{author}{Montesino-SanMartin, M.}, \bibinfo{author}{Militino, A.F.}, \bibinfo{author}{Wang, Q.}, \bibinfo{author}{Atkinson, P.M.}, \bibinfo{author}{Ugarte, M.D.}, \bibinfo{year}{2023}.
\newblock \bibinfo{title}{Unpaired spatio-temporal fusion of image patches (ustfip) from cloud covered images}.
\newblock \bibinfo{journal}{Remote Sensing of Environment} \bibinfo{volume}{295}, \bibinfo{pages}{113709}.
%Type = Article
\bibitem[{Heo and FitzHugh(2000)}]{heo2000standardized}
\bibinfo{author}{Heo, J.}, \bibinfo{author}{FitzHugh, T.W.}, \bibinfo{year}{2000}.
\newblock \bibinfo{title}{A standardized radiometric normalization method for change detection using remotely sensed imagery}.
\newblock \bibinfo{journal}{Photogrammetric Engineering and Remote Sensing} \bibinfo{volume}{66}, \bibinfo{pages}{173--181}.
%Type = Article
\bibitem[{Jiang et~al.(2022)Jiang, Shen and Li}]{Jiang2022STSf}
\bibinfo{author}{Jiang, M.}, \bibinfo{author}{Shen, H.}, \bibinfo{author}{Li, J.}, \bibinfo{year}{2022}.
\newblock \bibinfo{title}{Deep-learning-based spatio-temporal-spectral integrated fusion of heterogeneous remote sensing images}.
\newblock \bibinfo{journal}{IEEE Transactions on Geoscience and Remote Sensing} \bibinfo{volume}{60}, \bibinfo{pages}{1--15}.
\newblock \DOIprefix\doi{10.1109/TGRS.2022.3188998}.
%Type = Article
\bibitem[{Meng et~al.(2025)Meng, Chen, Zhang, Zhu, Zhang and Atkinson}]{MENG2025GLOSTFM}
\bibinfo{author}{Meng, Q.}, \bibinfo{author}{Chen, S.}, \bibinfo{author}{Zhang, L.}, \bibinfo{author}{Zhu, X.}, \bibinfo{author}{Zhang, Y.}, \bibinfo{author}{Atkinson, P.M.}, \bibinfo{year}{2025}.
\newblock \bibinfo{title}{Glostfm: A global spatiotemporal fusion model integrating multi-source satellite observations to enhance land surface temperature resolution}.
\newblock \bibinfo{journal}{Remote Sensing of Environment} \bibinfo{volume}{319}, \bibinfo{pages}{114640}.
\newblock \URLprefix \url{https://www.sciencedirect.com/science/article/pii/S0034425725000446}, \DOIprefix\doi{https://doi.org/10.1016/j.rse.2025.114640}.
%Type = Article
\bibitem[{Militino et~al.(2019)Militino, Ugarte, Pérez-Goya and Genton}]{militino2019ima}
\bibinfo{author}{Militino, A.F.}, \bibinfo{author}{Ugarte, M.D.}, \bibinfo{author}{Pérez-Goya, U.}, \bibinfo{author}{Genton, M.G.}, \bibinfo{year}{2019}.
\newblock \bibinfo{title}{Interpolation of the mean anomalies for cloud filling in land surface temperature and normalized difference vegetation index}.
\newblock \bibinfo{journal}{IEEE Transactions on Geoscience and Remote Sensing} \bibinfo{volume}{57}, \bibinfo{pages}{6068--6078}.
\newblock \DOIprefix\doi{10.1109/TGRS.2019.2904193}.
%Type = Article
\bibitem[{Moosavi et~al.(2015)Moosavi, Talebi, Mokhtari, Shamsi and Niazi}]{MOOSAVI2015NN}
\bibinfo{author}{Moosavi, V.}, \bibinfo{author}{Talebi, A.}, \bibinfo{author}{Mokhtari, M.H.}, \bibinfo{author}{Shamsi, S.R.F.}, \bibinfo{author}{Niazi, Y.}, \bibinfo{year}{2015}.
\newblock \bibinfo{title}{A wavelet-artificial intelligence fusion approach (waifa) for blending landsat and modis surface temperature}.
\newblock \bibinfo{journal}{Remote Sensing of Environment} \bibinfo{volume}{169}, \bibinfo{pages}{243--254}.
\newblock \URLprefix \url{https://www.sciencedirect.com/science/article/pii/S0034425715301036}, \DOIprefix\doi{https://doi.org/10.1016/j.rse.2015.08.015}.
%Type = Misc
\bibitem[{{ORNL DAAC}(2017)}]{modis}
\bibinfo{author}{{ORNL DAAC}}, \bibinfo{year}{2017}.
\newblock \bibinfo{title}{Modis collection 6 land product subsets web service}.
\newblock \URLprefix \url{https://daac.ornl.gov/cgi-bin/dsviewer.pl?ds_id=1557}, \DOIprefix\doi{10.3334/ORNLDAAC/1557}.
%Type = Article
\bibitem[{Pettorelli et~al.(2018)Pettorelli, Schulte~to B{\"u}hne, Tulloch, Dubois, Macinnis-Ng, Queir{\'o}s, Keith, Wegmann, Schrodt, Stellmes et~al.}]{ecology2018}
\bibinfo{author}{Pettorelli, N.}, \bibinfo{author}{Schulte~to B{\"u}hne, H.}, \bibinfo{author}{Tulloch, A.}, \bibinfo{author}{Dubois, G.}, \bibinfo{author}{Macinnis-Ng, C.}, \bibinfo{author}{Queir{\'o}s, A.M.}, \bibinfo{author}{Keith, D.A.}, \bibinfo{author}{Wegmann, M.}, \bibinfo{author}{Schrodt, F.}, \bibinfo{author}{Stellmes, M.}, et~al., \bibinfo{year}{2018}.
\newblock \bibinfo{title}{Satellite remote sensing of ecosystem functions: opportunities, challenges and way forward}.
\newblock \bibinfo{journal}{Remote Sensing in Ecology and Conservation} \bibinfo{volume}{4}, \bibinfo{pages}{71--93}.
%Type = Article
\bibitem[{Pons et~al.(2014)Pons, Pesquer, Crist{\'o}bal and Gonz{\'a}lez-Guerrero}]{pons2014automatic}
\bibinfo{author}{Pons, X.}, \bibinfo{author}{Pesquer, L.}, \bibinfo{author}{Crist{\'o}bal, J.}, \bibinfo{author}{Gonz{\'a}lez-Guerrero, O.}, \bibinfo{year}{2014}.
\newblock \bibinfo{title}{Automatic and improved radiometric correction of landsat imagery using reference values from modis surface reflectance images}.
\newblock \bibinfo{journal}{International Journal of Applied Earth Observation and Geoinformation} \bibinfo{volume}{33}, \bibinfo{pages}{243--254}.
%Type = Article
\bibitem[{Roy et~al.(2014)Roy, Wulder, Loveland, Woodcock, Allen, Anderson, Helder, Irons, Johnson, Kennedy et~al.}]{landsat8}
\bibinfo{author}{Roy, D.P.}, \bibinfo{author}{Wulder, M.A.}, \bibinfo{author}{Loveland, T.R.}, \bibinfo{author}{Woodcock, C.}, \bibinfo{author}{Allen, R.G.}, \bibinfo{author}{Anderson, M.C.}, \bibinfo{author}{Helder, D.}, \bibinfo{author}{Irons, J.R.}, \bibinfo{author}{Johnson, D.M.}, \bibinfo{author}{Kennedy, R.}, et~al., \bibinfo{year}{2014}.
\newblock \bibinfo{title}{Landsat-8: Science and product vision for terrestrial global change research}.
\newblock \bibinfo{journal}{Remote sensing of Environment} \bibinfo{volume}{145}, \bibinfo{pages}{154--172}.
%Type = Article
\bibitem[{Shao et~al.(2019)Shao, Cai, Fu, Hu and Liu}]{SHAO2019DLfusion}
\bibinfo{author}{Shao, Z.}, \bibinfo{author}{Cai, J.}, \bibinfo{author}{Fu, P.}, \bibinfo{author}{Hu, L.}, \bibinfo{author}{Liu, T.}, \bibinfo{year}{2019}.
\newblock \bibinfo{title}{Deep learning-based fusion of landsat-8 and sentinel-2 images for a harmonized surface reflectance product}.
\newblock \bibinfo{journal}{Remote Sensing of Environment} \bibinfo{volume}{235}, \bibinfo{pages}{111425}.
\newblock \URLprefix \url{https://www.sciencedirect.com/science/article/pii/S0034425719304444}, \DOIprefix\doi{https://doi.org/10.1016/j.rse.2019.111425}.
%Type = Article
\bibitem[{Shen(2012)}]{Shen2012STSf}
\bibinfo{author}{Shen, H.}, \bibinfo{year}{2012}.
\newblock \bibinfo{title}{Integrated fusion method for multiple temporal-spatial-spectral images}.
\newblock \bibinfo{journal}{The International Archives of the Photogrammetry, Remote Sensing and Spatial Information Sciences} \bibinfo{volume}{XXXIX-B7}, \bibinfo{pages}{407--410}.
\newblock \URLprefix \url{https://isprs-archives.copernicus.org/articles/XXXIX-B7/407/2012/}, \DOIprefix\doi{10.5194/isprsarchives-XXXIX-B7-407-2012}.
%Type = Article
\bibitem[{Sommervold et~al.(2023)Sommervold, Gazzea and Arghandeh}]{sommervold2023survey}
\bibinfo{author}{Sommervold, O.}, \bibinfo{author}{Gazzea, M.}, \bibinfo{author}{Arghandeh, R.}, \bibinfo{year}{2023}.
\newblock \bibinfo{title}{A survey on sar and optical satellite image registration}.
\newblock \bibinfo{journal}{Remote Sensing} \bibinfo{volume}{15}, \bibinfo{pages}{850}.
%Type = Article
\bibitem[{Tang et~al.(2020)Tang, Wang, Zhang and Atkinson}]{coreg2020}
\bibinfo{author}{Tang, Y.}, \bibinfo{author}{Wang, Q.}, \bibinfo{author}{Zhang, K.}, \bibinfo{author}{Atkinson, P.M.}, \bibinfo{year}{2020}.
\newblock \bibinfo{title}{Quantifying the effect of registration error on spatio-temporal fusion}.
\newblock \bibinfo{journal}{IEEE Journal of Selected Topics in Applied Earth Observations and Remote Sensing} \bibinfo{volume}{13}, \bibinfo{pages}{487--503}.
%Type = Article
\bibitem[{Vivone et~al.(2025)Vivone, Deng, Deng, Hong, Jiang, Li, Li, Shen, Wu, Xiao, Yao, Zhang, Chanussot, García and Plaza}]{vivone2025DLreview}
\bibinfo{author}{Vivone, G.}, \bibinfo{author}{Deng, L.J.}, \bibinfo{author}{Deng, S.}, \bibinfo{author}{Hong, D.}, \bibinfo{author}{Jiang, M.}, \bibinfo{author}{Li, C.}, \bibinfo{author}{Li, W.}, \bibinfo{author}{Shen, H.}, \bibinfo{author}{Wu, X.}, \bibinfo{author}{Xiao, J.L.}, \bibinfo{author}{Yao, J.}, \bibinfo{author}{Zhang, M.}, \bibinfo{author}{Chanussot, J.}, \bibinfo{author}{García, S.}, \bibinfo{author}{Plaza, A.}, \bibinfo{year}{2025}.
\newblock \bibinfo{title}{Deep learning in remote sensing image fusion: Methods, protocols, data, and future perspectives}.
\newblock \bibinfo{journal}{IEEE Geoscience and Remote Sensing Magazine} \bibinfo{volume}{13}, \bibinfo{pages}{269--310}.
\newblock \DOIprefix\doi{10.1109/MGRS.2024.3495516}.
%Type = Article
\bibitem[{Wang et~al.(2020)Wang, Wang, Wang and Atkinson}]{coreg2020patch}
\bibinfo{author}{Wang, L.}, \bibinfo{author}{Wang, X.}, \bibinfo{author}{Wang, Q.}, \bibinfo{author}{Atkinson, P.M.}, \bibinfo{year}{2020}.
\newblock \bibinfo{title}{Investigating the influence of registration errors on the patch-based spatio-temporal fusion method}.
\newblock \bibinfo{journal}{IEEE Journal of Selected Topics in Applied Earth Observations and Remote Sensing} \bibinfo{volume}{13}, \bibinfo{pages}{6291--6307}.
%Type = Article
\bibitem[{Wang and Atkinson(2018)}]{fitfc2018}
\bibinfo{author}{Wang, Q.}, \bibinfo{author}{Atkinson, P.M.}, \bibinfo{year}{2018}.
\newblock \bibinfo{title}{Spatio-temporal fusion for daily sentinel-2 images}.
\newblock \bibinfo{journal}{Remote Sensing of Environment} \bibinfo{volume}{204}, \bibinfo{pages}{31--42}.
%Type = Article
\bibitem[{Weiss et~al.(2020)Weiss, Jacob and Duveiller}]{agro2020}
\bibinfo{author}{Weiss, M.}, \bibinfo{author}{Jacob, F.}, \bibinfo{author}{Duveiller, G.}, \bibinfo{year}{2020}.
\newblock \bibinfo{title}{Remote sensing for agricultural applications: A meta-review}.
\newblock \bibinfo{journal}{Remote Sensing of Environment} \bibinfo{volume}{236}, \bibinfo{pages}{111402}.
%Type = Article
\bibitem[{Wu et~al.(2021)Wu, Liu, Zhu, Bai, Miao, Ma and Gong}]{wu2021computational}
\bibinfo{author}{Wu, Y.}, \bibinfo{author}{Liu, J.W.}, \bibinfo{author}{Zhu, C.Z.}, \bibinfo{author}{Bai, Z.F.}, \bibinfo{author}{Miao, Q.G.}, \bibinfo{author}{Ma, W.P.}, \bibinfo{author}{Gong, M.G.}, \bibinfo{year}{2021}.
\newblock \bibinfo{title}{Computational intelligence in remote sensing image registration: A survey}.
\newblock \bibinfo{journal}{International Journal of Automation and Computing} \bibinfo{volume}{18}, \bibinfo{pages}{1--17}.
%Type = Misc
\bibitem[{Zhao and Wentz(2020)}]{urban2020}
\bibinfo{author}{Zhao, Q.}, \bibinfo{author}{Wentz, E.A.}, \bibinfo{year}{2020}.
\newblock \bibinfo{title}{Editorial for the special issue:“remote sensing of urban ecology and sustainability”}.
%Type = Article
\bibitem[{Zhou et~al.(2021)Zhou, Chen, Chen, Zhu, Qiu, Song, Rao, Zhang, Cao and Cui}]{zhou2021}
\bibinfo{author}{Zhou, J.}, \bibinfo{author}{Chen, J.}, \bibinfo{author}{Chen, X.}, \bibinfo{author}{Zhu, X.}, \bibinfo{author}{Qiu, Y.}, \bibinfo{author}{Song, H.}, \bibinfo{author}{Rao, Y.}, \bibinfo{author}{Zhang, C.}, \bibinfo{author}{Cao, X.}, \bibinfo{author}{Cui, X.}, \bibinfo{year}{2021}.
\newblock \bibinfo{title}{Sensitivity of six typical spatiotemporal fusion methods to different influential factors: A comparative study for a normalized difference vegetation index time series reconstruction}.
\newblock \bibinfo{journal}{Remote Sensing of Environment} \bibinfo{volume}{252}, \bibinfo{pages}{112130}.
%Type = Article
\bibitem[{Zhu et~al.(2022)Zhu, Zhan, Zhou, Chen, Liang, Xu and Chen}]{zhu2022novel}
\bibinfo{author}{Zhu, X.}, \bibinfo{author}{Zhan, W.}, \bibinfo{author}{Zhou, J.}, \bibinfo{author}{Chen, X.}, \bibinfo{author}{Liang, Z.}, \bibinfo{author}{Xu, S.}, \bibinfo{author}{Chen, J.}, \bibinfo{year}{2022}.
\newblock \bibinfo{title}{A novel framework to assess all-round performances of spatiotemporal fusion models}.
\newblock \bibinfo{journal}{Remote Sensing of Environment} \bibinfo{volume}{274}, \bibinfo{pages}{113002}.

\end{thebibliography}

\end{document}